\documentclass[conference]{IEEEtran}
\IEEEoverridecommandlockouts
\usepackage{cite}
\usepackage{amsmath,amssymb,amsfonts}
\usepackage{graphicx}
\usepackage{textcomp}
\usepackage{xcolor}
\usepackage{soul}
\usepackage{listings}
\usepackage{acronym}
\usepackage[linesnumbered,ruled,vlined]{algorithm2e}
\usepackage{subcaption}
\usepackage{import}
\usepackage[hyphens]{url}
\usepackage[hidelinks]{hyperref}
\hypersetup{breaklinks=true}
\usepackage{cleveref}
\acrodef{FL}{federated learning}
\newcommand{\approach}{PSFL}
\graphicspath{{figures/}}

\def\BibTeX{{\rm B\kern-.05em{\sc i\kern-.025em b}\kern-.08em
    T\kern-.1667em\lower.7ex\hbox{E}\kern-.125emX}}

\lstdefinelanguage{scala}{
    keywords={abstract,case,catch,class,def,%
        do,else,extends,false,final,finally,%
        for,if,implicit,import,match,mixin,%
        new,null,object,override,package,%
        private,protected,requires,return,sealed,%
        super,this,throw,trait,true,try,lazy,%
        type,val,var,while,with,yield,forSome},
    otherkeywords={=>,<-,<\%,<:,>:,\#},
    sensitive=true,
    columns=fullflexible,
    morecomment=[l]{//},
    morecomment=[n]{/*}{*/},
    morestring=[b]",
    stringstyle=\ttfamily\color{red!50!brown},
    showstringspaces=false,
    morestring=[eratb]',
    morestring=[b]""",
    basicstyle=\sffamily\lst@ifdisplaystyle\footnotesize\fi\ttfamily,
    emphstyle=\sffamily\bfseries\ttfamily
    }
    \definecolor{ddarkgreen}{rgb}{0,0.5,0}
    \lstdefinelanguage{scafi}{
    frame=single,
    basewidth=0.5em,
    language={scala},
    keywordstyle=\color{blue}\textbf,
    commentstyle=\color{ddarkgreen},
    keywordstyle=[2]\color{teal}\textbf,
    keywords=[2]{rep,nbr,foldhood,foldhoodPlus,aggregate,branch,spawn,mux,mid},
    keywordstyle=[3]\color{gray},
    keywords=[3]{Me,AroundMe,Everywhere,Forever}, %,@@,@@@
    keywordstyle=[4]\color{red}\textbf,
    keywords=[4]{in,out,rd},
    keywordstyle=[5]\color{violet},
    keywords=[5]{evolve,when,andNext,workflow,G,C,broadcast,gossip, CWithSenderField, classicGradient},
    keywordstyle=[6]\color{orange},
    keywords=[6]{Available,Serving,Done,Waiting,Removing,None,Set}
    }

\begin{document}

\title{Proximity-based Self-Federated Learning\\
{}
%\thanks{Identify applicable funding agency here. If none, delete this.}
}

\author{
%\IEEEauthorblockN{Anonymous Authors}}
%\IEEEauthorblockA{
%	\textit{University}\\
%	Country\\
%	email
%	}
\IEEEauthorblockN{Davide Domini}
\IEEEauthorblockA{
%\textit{Department of Computer Science and Engineering} \\
\textit{%Alma Mater Studiorum--
University of Bologna}\\
Cesena, Italy\\
davide.domini@unibo.it
}
\and 
\IEEEauthorblockN{Gianluca Aguzzi} %\IEEEauthorrefmark{1}
\IEEEauthorblockA{
%\textit{Department of Computer Science and Engineering} \\
\textit{%Alma Mater Studiorum--
University of Bologna}\\
Cesena, Italy\\
gianluca.aguzzi@unibo.it % 0000−0001−9149−949X
}
\and
\IEEEauthorblockN{Nicolas Farabegoli} %\IEEEauthorrefmark{1}
\IEEEauthorblockA{
%\textit{Department of Computer Science and Engineering} \\
\textit{%Alma Mater Studiorum--
University of Bologna}\\
Cesena, Italy\\
nicolas.farabegoli@unibo.it % 0000−0001−9149−949X
}
\and
%\linebreakand %\and
\IEEEauthorblockN{Mirko Viroli}
\IEEEauthorblockA{
%\textit{Department of Computer Science and Engineering} \\
\textit{%Alma Mater Studiorum--
University of Bologna}\\
Cesena, Italy\\
mirko.viroli@unibo.it % 0000−0003−2702−5702
}
\and
%\linebreakand %\and
\IEEEauthorblockN{Lukas Esterle}
\IEEEauthorblockA{
	%\textit{Department of Computer Science and Engineering} \\
	\textit{%Alma Mater Studiorum--
		Aarhus University}\\
	Aarhus, Denmark\\
	lukas.esterle@ece.au.dk % 0000−0003−2702−5702
}
}
%meta command
\newcommand{\meta}[1]{\textcolor{blue}{#1}}
\newcommand{\lukas}[1]{\textcolor{red}{\small #1}}
\newcommand{\gianluca}[1]{\textcolor{green}{\small #1}}

\maketitle

\begin{abstract}
In recent advancements in machine learning, 
 \acl{FL} allows a network of distributed clients 
 to collaboratively develop a global model without needing to share their local data. 
This technique aims to safeguard privacy, 
 countering the vulnerabilities of conventional centralized learning methods. 
 Traditional \acl{FL} approaches often rely on a central server to coordinate model training across clients,
 aiming to replicate the same model uniformly across all nodes.
However, these methods overlook the significance of geographical and local data variances in vast networks, 
 potentially affecting model effectiveness and applicability.
Moreover, relying on a central server might become a bottleneck in large networks, 
 such as the ones promoted by edge computing.
Our paper introduces a novel, fully-distributed \acl{FL} strategy called \emph{proximity-based self-federated learning} 
 that enables the self-organised creation of multiple \emph{federations} of clients based on their geographic proximity and data distribution without exchanging raw data.
 Indeed, unlike traditional algorithms, our approach encourages clients to share and adjust their models with neighbouring nodes based on geographic proximity and model accuracy. 
 This method not only addresses the limitations posed by diverse data distributions but also enhances the model's adaptability to different regional characteristics creating specialized models for each federation. 
We demonstrate the efficacy of our approach through simulations on well-known datasets, showcasing its effectiveness over the conventional centralized federated learning framework.
\end{abstract}
\section{Introduction}

\Ac{FL}~\cite{DBLP:conf/aistats/McMahanMRHA17,DBLP:journals/corr/abs-2301-01299} has dramatically transformed the landscape of machine learning, 
enabling collaborative model training across geographically distributed datasets while \emph{upholding} data privacy. 
This paradigm is particularly crucial in scenarios where data contains sensitive information, 
such as healthcare or personal mobile device usage~\cite{DBLP:conf/iccnc/ChenXKWH0H20, DBLP:journals/toit/PfitznerSA21}, 
which are legally or ethically restricted from central aggregation.

%\gianluca{Lukas comment: your new introduction reads really nice but i am missing the link from being centralized to distributed and the neglect of specialised models. i will add a sentence or two in the second paragraph, preparing the reader for the contribution in the third paragraph...}
Traditional \ac{FL} employs a central server to coordinate the aggregation of model updates across a network, 
 which, while effective in smaller, controlled environments, 
 scales poorly with the expansion of network size and geographic distribution. 
 This central reliance becomes a bottleneck, introducing significant delays and increased communication overhead, 
 which are further exacerbated by the growth in edge computing and IoT devices~\cite{DBLP:journals/fgcs/OlaniyanFMZ18}. 
Recent \ac{FL} adaptations, such as hierarchical and distributed approaches~\cite{DBLP:conf/icc/Liu0SL20, DBLP:journals/corr/abs-2203-12281}, reduce the reliance of single central components to manage the federation process.
 However, these approaches still integrate all models equally and therefore fail to account for the variations in data distribution across different regions~\cite{esterle2022deep}. This leads to poor generalization across different geographic regions and underlying operational contexts and can adversely affect the model's effectiveness and accuracy.
Addressing these limitations, 
 this paper introduces a novel concept in federated learning called \emph{proximity-based self-federated learning} (\approach{}). 
 Our methodology diverges from traditional \acl{FL} by promoting a \emph{decentralized federation} of models where nodes form clusters or federations based on both \emph{geographic proximity}~\cite{DBLP:conf/coordination/DominiAEV24} and \emph{data similarity}. 
 By leveraging principles from \emph{self-organizing} systems and inspired by novel \emph{macroprogramming} paradigms (i.e., programming paradigms which have a collective abstraction as a first-class citizen~\cite{DBLP:journals/csur/Casadei23}), like \emph{aggregate computing}~\cite{DBLP:journals/computer/BealPV15} (i.e., a top-down global-to-local programming model which express collective computation as a composition of computational fields), 
 \approach{} allows nodes within close physical or network proximity to collaborate more closely on model training, dynamically causing the systems to partition into federations where a single shared model emerges,
 thus enhancing the relevance and effectiveness of the federation-wise learned models.

The main contributions of this paper are:
\begin{itemize}
    \item We propose a new decentralized \acl{FL} framework that enables dynamic federation formation based on geographic and data-driven criteria based on \emph{space-fluid computing}~\cite{DBLP:journals/lmcs/CasadeiMPVZ23} (i.e., a modern approach to sample a spatially distributed phenomena) without the need of sharing data and a central coordinating node.
    %\item We incorporate the concept of \emph{space-fluid computing} to %dynamically adapt the federations' definition to the geographic distribution of data, thus aligning model training more closely with the underlying data characteristics.
    \item Through simulations on well-known datasets, we demonstrate that \approach{} significantly outperforms traditional centralized \acl{FL} approaches, 
    particularly in largely heterogeneous environments.
\end{itemize}
The rest of the paper is structured as follows: \Cref{sec:background} provides the context in which our work is situated, \Cref{sec:approach} details the proposed \approach{} approach and its implementation, \Cref{sec:evaluation} presents the evaluation results, and \Cref{sec:conclusion} concludes the paper with a discussion of future work.
%By fundamentally rethinking the coordination in \acl{FL}, \approach{} offers a robust solution to the challenges posed by the next generation of distributed systems, paving the way for more efficient and context-aware machine learning models.
% TODO add more details from there
%\meta{Section length 1 page + abstract}
\section{Background and Related Works}\label{sec:background}
\subsection{Self-organising Coordination Region}\label{sec:scr}
This paper addresses the challenges posed by performing distributed cooperative learning in modern highly distributed systems---including pervasive computing, collective adaptive systems, the internet of things, cyber-physical systems, and edge computing.
 These systems exhibit distinctive characteristics that complicate their management like \emph{distribution}, \emph{situatedness}, and \emph{scale}, and they pose significant challenges in coordinating these systems~\cite{DBLP:conf/icdcs/AguzziCV22,DBLP:conf/coordination/CasadeiPVN19}, including: 
\begin{enumerate}
    %\item \emph{Task complexity and collaboration:} heterogeneity necessitates collaboration due to asymmetries in component capabilities for handling complex tasks.
    \item \emph{Locality principle:} operational efficiency and cost are influenced by the principle of locality, with dependencies on the proximity of sources, processes, and users.
    \item \emph{Control and decision-making:} a balance between centralized and decentralized control is essential, as extremes are neither feasible nor desirable.
    \item \emph{Dynamic environments:} constant changes in the environment and system structure, driven by factors such as mobility and component failures, demand adaptive responses.
\end{enumerate}
A recurrent solution in such scenarios consists in \emph{dynamically} partitioning large network in contiguous regions such that the nodes in each region can efficiently cooperate one another.
 This approach is particularly effective in scenarios where the data distribution is highly dynamic and can change rapidly over time. 
 These regions can be used for several purposes, 
 from decentralized services orchestration to enhance scalability and performance~\cite{DBLP:journals/concurrency/JaradatDB16}, 
 WSN middleware for organized communication and power management~\cite{DBLP:conf/icw/DiazRT05}, 
 distributed data processing in IoT networks~\cite{DBLP:journals/internet/AguzziCPV22},
 and dynamic control in robot swarms~\cite{DBLP:journals/corr/abs-2401-10969}.

An example solution to achieve such partitioning without global supervisioning is the \emph{self-organizing coordination region} (SCR)~\cite{DBLP:conf/coordination/CasadeiPVN19} pattern, 
 which works as follows: 
 a system-wide \emph{multi-leader election} process determines a set of leaders among a set of leader candidates; 
 the system (or its environment) forms a self-organising set of regions, 
 each one regulated by a \emph{single} leader; 
 within each region, a \emph{feedback loop} is established, 
 whereby the leader receives \emph{upstream information flows} from the members of the region (possibly leveraging intermediaries) and emits a \emph{control information flow downstream}.

Information within these regions is disseminated primarily through \emph{gossip protocols} and \emph{gradient-based information casts}~\cite{DBLP:conf/saso/AudritoCDV17}. 
 Gossip protocols are effective for disseminating information with the cost of a higher number of messages,
 while gradient-based approaches propagate information across increasing distances, useful for defining regional boundaries and leader-user communication. 
 
Accumulating information, especially in larger networks, poses greater challenges. 
 While gossip methods can suffice in small regions, 
 scalable solutions like \emph{spanning tree} and \emph{multi-path techniques} are necessary for larger areas. 
 %Spanning trees, although prone to disruptions due to network changes, 
 %are typically first-line methods, whereas multi-path approaches provide redundancy and improve robustness but may slow convergence in stable networks.

SCR can be implemented using various programming paradigms, 
 but it is particularly effective when using \emph{aggregate computing}~\cite{DBLP:journals/computer/BealPV15}, which provides tools for expressing self-organising computations via functional composition of smaller ``bricks'' enhancing modularity and reusability. 
 These ``bricks'' are:
\begin{itemize}
    \item S (for Sparse-choice—i.e., a scattered selection from the set of participating devices) and its variants, like the one based on \emph{space-fluid sampling}~\cite{DBLP:journals/lmcs/CasadeiMPVZ23} (i.e., the leader are selected based on a perception of a spatial distributed phenomena),
    \item G (for Gradient-cast—i.e., a multicast diffusing information along a gradient),
    \item C (for Converge-cast—i.e., a multicast aggregating information to a sink device).
\end{itemize}
Moreover, this paradigm seems to be particularly effective in the context of distributed machine learning, 
 as already demonstrated in related works~\cite{DBLP:conf/acsos/AguzziVE23,DBLP:conf/acsos/AguzziCV22,DBLP:conf/coordination/AguzziCV22,DBLP:conf/icdcs/AguzziCV22}.
\subsection{Federated Learning}

Federated learning~\cite{DBLP:conf/aistats/McMahanMRHA17,DBLP:journals/corr/abs-2301-01299} is a machine learning framework that aims to collaboratively train a unique global model from distributed datasets. 
    This technique has been introduced to enable training models without collecting and merging various datasets into a single central server, thus making it possible to work
    in contexts with strong privacy concerns (e.g., hospital or banking data).
    Furthermore, in highly distributed systems, given the large amount of data that can be generated by various clients, it becomes infeasible 
    to move all data to a single central server~\cite{DBLP:journals/comsur/NguyenDPSLP21}. In this context, \ac{FL} becomes crucial as it allows leveraging the computational capabilities of the various clients.

%\ac{FL} can be categorized into three macro-categories depending on how data are partitioned among various devices~\cite{DBLP:journals/kbs/ZhangXBYLG21}
%:
%    i) \emph{Horizontal Federated Learning}: each device shares the same feature space while possessing different samples,
%    ii) \emph{Vertical Federated Learning}: devices have different feature spaces while their samples overlap, and
%    iii) \emph{Federated Transfer Learning}: aims to extend knowledge from a specific application domain to another domain characterized by sparse data.

\begin{figure}
    % \centerline{\includegraphics[width=0.4\textwidth]{figures/federated-learning-schema.pdf}}
    \centering
    \def\svgwidth{0.4\textwidth}
    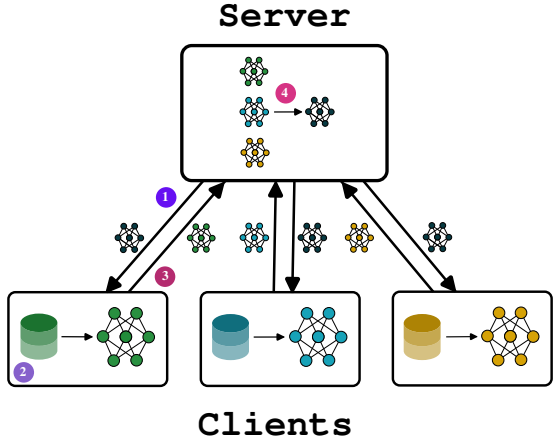
    \caption{Federated learning schema. 
    In the first phase, the server shares the centralized model with the clients. 
    In the second phase, 
    the clients perform a local learning phase using data that is not accessible to the server. 
    In the third phase, these models are communicated back to the central server, and finally, in the last phase, there is an aggregation algorithm. 
    }
    \label{fig:fl-schema}
\end{figure}

    Even though various versions of FL exist~\cite{DBLP:journals/kbs/ZhangXBYLG21}, they all share a common learning flow. 
    Traditionally (as shown in \Cref{fig:fl-schema}), \ac{FL} is based on a client-server architecture 
    and comprises the following steps:
    \begin{enumerate}
        \item \emph{Model initialization}: the central server initializes a common base model that is shared with each client;
        \item \emph{Local learning}: each client performs one or more steps of local learning on its own dataset;
        \item \emph{Local models sharing}: each client sends back to the central server the new model trained on its own data (i.e. the local model);
        \item \emph{Local models aggregation}: the local models collected by the central server are aggregated to obtain the new global model.
    \end{enumerate}
    This process is carried out iteratively for a predefined number of global rounds.

%Although it is a very promising technique, 
The classical client-server architecture presents some limitations, namely:
    i) in federations with many clients, the server may be a bottleneck;
    ii) the server is a single point of failure, and if it fails to communicate with the clients, the entire the learning process is interrupted; and
    iii) the server must be a trusted entity, which could be a challenging constraint in some scenarios.
    For these reasons, various distributed federated learning approaches have been introduced~\cite{DBLP:journals/jpdc/HegedusDJ21,DBLP:conf/dsn/WinkN21,DBLP:journals/tvt/GuptaLCCN24}, based on peer-to-peer or semi-centralized architectures. 

\subsection{Data heterogeneity}
\label{sec:data-het}
Federated learning algorithms achieve remarkable performance when data are homogeneously distributed among clients. However, in real-life datasets, 
    data are often heterogeneous, namely \emph{non-independently and identically distributed} (Non-IID). 
    This heterogeneity can be caused by various factors~\cite{DBLP:journals/ijon/ZhuXLJ21}. 
    For instance, consider a scenario in which we have a huge amount of stations for PM10 sampling distributed across Europe.
    %We aim to perform an air quality classification task using federated learning.
    In this case, data heterogeneity emerges since stations grouped in a certain geographical area have similar data distributions among themselves but different from those in other areas.
    % For instance, consider a task where we aim to perform federated learning across different
    % smartphones for text generation. It is common for clients grouped in a certain geographical area to use similar words, expressions, 
    % and emojis among themselves but different from those in other areas.
    Data heterogeneity can be categorized in various ways based on how the data are distributed among the clients~\cite{DBLP:conf/icde/LiDCH22,DBLP:journals/ftml/KairouzMABBBBCC21}. 
    The main categories include:
    i) \emph{feature skew}: all clients have the same labels but different feature distributions (e.g., in a handwritten text classification task, we may have the same letters written in different calligraphic styles);
    ii) \emph{label skew}: each client has only a subset of the classes; and
    iii) \emph{quantity skew}: each client has a significantly different amount of data compared to others.

Non-IID data is a critical aspect because in such scenarios, individual clients have local objectives that diverge from the global one~\cite{DBLP:conf/icde/LiDCH22}. As a result, after the local training, 
    the clients produce vastly different updates (i.e., \emph{local update drift}) that, once aggregated, lead to a reduction in the accuracy of the global model or 
    convergence issues~\cite{DBLP:journals/ftml/KairouzMABBBBCC21}. 
    For these reasons, \ac{FL} with non-IID data is a growing research field. In the literature, there are various families of approaches aiming to enhance learning in these scenarios. 
    One approach involves attempting to train a single global model by improving base algorithms (e.g., FedAvg~\cite{DBLP:journals/corr/McMahanMRA16}) through the addition of regularization terms to constrain the local updates or 
    considering that each device may perform a different number of local training steps. 
    This is implemented by various algorithms, including: FedProx~\cite{DBLP:conf/mlsys/LiSZSTS20}, 
    Scaffold~\cite{DBLP:journals/corr/abs-1910-06378}, and FedNova~\cite{DBLP:conf/nips/WangLLJP20}.
    Another possible approach -- the one we focus on -- involves dividing the devices into clusters and having multiple global models, one for each cluster, in order 
    to create specialized models.

\subsection{Clustered Federated Learning}

Clustered \acl{FL} is an approach to tackle data heterogeneity. 
It is a technique of personalized \ac{FL} in which, instead of a single global model, \emph{multiple} models are trained to target various
    local distributions. It is based on the assumption that clients can be divided into subgroups (i.e., \emph{clusters}), 
    and that within each cluster, clients have data that follow similar distributions (i.e., \emph{IID} data).
    Moreover, studies \cite{DBLP:journals/tit/GhoshCYR22, DBLP:journals/corr/abs-1910-01991} show that aggregating models among clients within the same cluster leads to improved performance and 
    personalization; while %, as also mentioned in \Cref{sec:data-het}, 
    aggregating models from different clusters results in a degradation of the accuracy of the global model. 
A key aspect of these approaches is how to measure the \emph{similarity} among various clients. Generally, there are three methods~\cite{9954190}, namely:
    i) \emph{loss-based}: clients measure the losses of the models from various clusters on their own data, with the model having the lowest loss considered most similar~\cite{DBLP:journals/tit/GhoshCYR22};
    ii) \emph{gradient-based}: clients measure similarity based on the distance between gradient updates~\cite{DBLP:journals/corr/abs-1910-01991, DBLP:journals/tpds/DuanLJWLCTR22, DBLP:journals/corr/abs-2010-06870}; and
    iii) \emph{weight-based}: clients measure similarity based on the weights of their models~\cite{DBLP:journals/corr/abs-1910-01991,DBLP:journals/tnn/NguyenPTHTSH23}.

In literature, several algorithms based on client clustering have been proposed, such as: PANM~\cite{9954190}, FedSKA~\cite{DBLP:conf/ecai/Li0TWXZ23}, IFCA~\cite{DBLP:journals/tit/GhoshCYR22}. 
    However, all these algorithms are based on centralized architectures and require prior knowledge of the number of clusters. This significantly impacts the final accuracy, 
    as an incorrect choice of this hyperparameter can lead to a considerable degradation in performance.

\section{Proximity-based self-federated learning}\label{sec:approach}
\begin{table}[ht]
    \centering
    \caption{Summary of Symbols in \approach{}}
    \begin{tabular}{@{}ll@{}}
    \hline
    \textbf{Symbol} & \textbf{Description} \\ \hline
    $A$ & Spatial area divided into $k$ distinct continuous areas \\ \hline
    $a_j$ & A specific region within $A$  \\ \hline
    $\Theta_i$ & Data distributions for areas $i$ \\ \hline
    $d'$ & Data samples from distribution $\Theta$ \\ \hline
    $m(d', d'')$ & Metric determining the disparity between two data \\ \hline
    $\delta$ & Error bound for dissimilarity intra-area and inter-area \\ \hline
    $S$ & Set of sensor nodes deployed in the area $A$ \\ \hline
    $s_i$ & A sensor node, part of the set $S$ \\ \hline
    %$p_i$ & Position of sensor $s_i$ \\ \hline
    $r_c$ & Communication range of a sensor node \\ \hline
    $N_i$ & Neighbourhood of node $i$ \\ \hline
    $D_i$ & Local dataset created by node $i$\\ \hline
    $nn_i^t$ & Local model of node $i$ at time step $t$ \\ \hline
    $T$ & Total number of time steps in the learning process \\ \hline
    $F^t$ & Set of federations at time step $t$ \\ \hline
    $f_j^t$ & A specific federation at time step $t$ \\ \hline
    $l_j$ & Leader node of federation $f_j$ \\ \hline
    $nn^T$ & Set of all federation-wise models at final time step $T$ \\ \hline
    $L(\Theta_i, nn_i^T)$ & Average error of model $nn_i^T$ on data distribution $\Theta_i$ \\ \hline
    $ds(i, j)$ & Dissimilarity measure based on loss functions $L_{i,j}$ and $L_{j,i}$ \\ \hline
    $G$ & Gradient field indicating accumulated error to a $l_i$ \\ \hline
    $\sigma$ & Maximum tolerable path error for federation coherence \\ \hline
    \end{tabular}
\end{table}
\begin{figure*}[htb]
    \centering
    \begin{subfigure}[b]{0.3\textwidth}
        \centering
        \includegraphics[width=\textwidth]{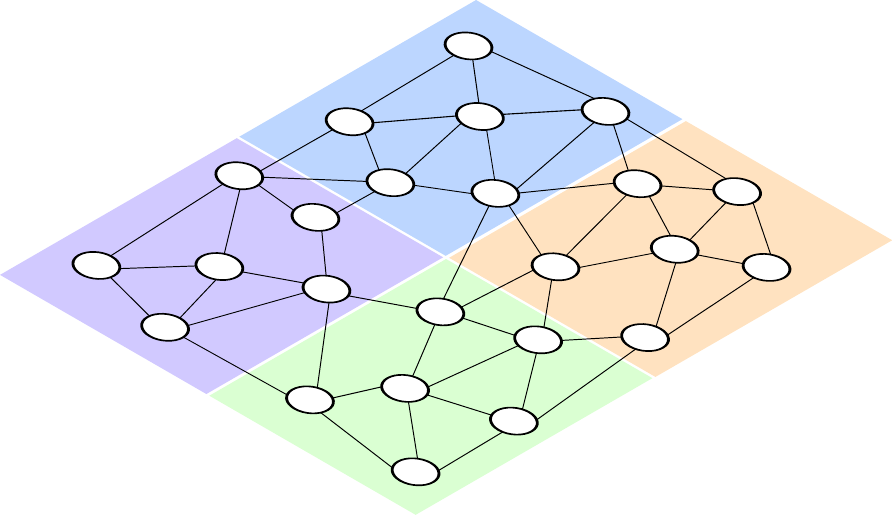}
        %\caption{Start of the simulation.}
        \label{fig:zones1}
    \end{subfigure}
    \begin{subfigure}[b]{0.3\textwidth}
        \centering
        \includegraphics[width=\textwidth]{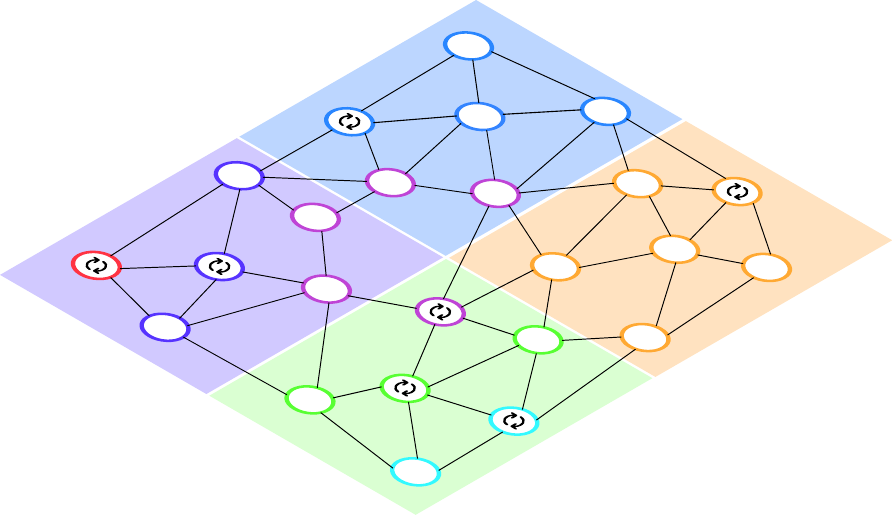}
        %\caption{After k time steps of the simulation.}
        \label{fig:zones2}
    \end{subfigure}
    \begin{subfigure}[b]{0.3\textwidth}
        \centering
        \includegraphics[width=\textwidth]{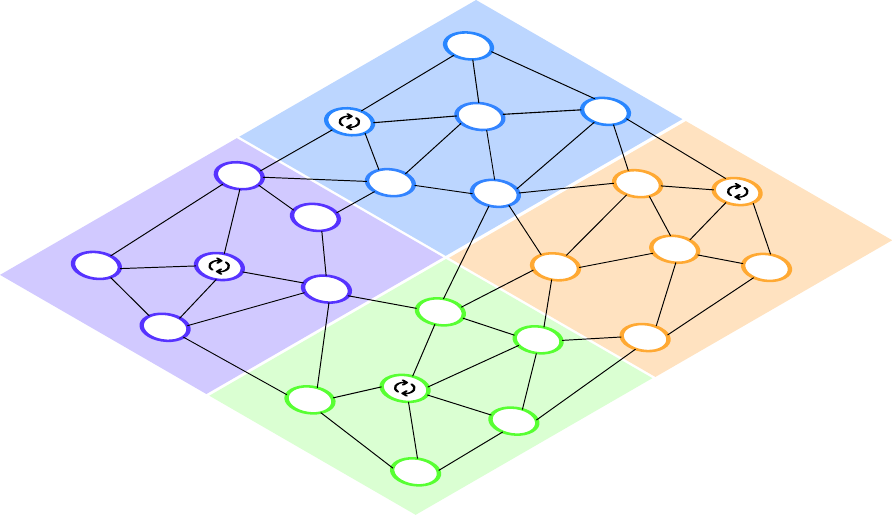}
        %\caption{End of the simulation.}
        \label{fig:zones3}
    \end{subfigure}
    \caption{Graphical representation of the problem.
    In this case, there is an area $A$ divided into four sub-areas $\{a_1, a_2, a_3, a_4\}$. 
    Each sub-area has a different data distribution (represented by the colours). 
    The white nodes represent sensors $S$, while links between sensors represent
    the neighbourhood relation. %(i.e., the ability of two sensors $s_i$ and $s_j$ to exchange information). 
    During the learning process (time flows from top to bottom), the nodes begin to form various federations 
    (represented by the colour of the nodes, while the leaders of the federations are represented by arrows), 
    eventually reaching a stable division that matches the areas they belong to.
    }
    \label{fig:zones-evolution}
\end{figure*}
\subsection{Problem Statement}
% We consider, as shown in \Cref{fig:zones-evolution}, a spatial area $A$ %with dimensions characterized by a width of $w_A$ and a height of $h_A$. 
%  %The area $A$ is 
%  divided into $k$ distinct \emph{continuous} areas of any shapes such that $A = \{a_1, a_2, \ldots, a_k\}$ represents the spatial area $A$. 
%  Each region $a_j$ has a unique local data distribution $\Theta_j$. %and provides specific localized information.
% %
As depicted in~\Cref{fig:zones-evolution},
we consider $A = \{a_1, a_2, \ldots, a_k\}$ the spatial area divided into $k$ distinct continuous areas.
Each area $a_j$ has a unique local data distribution $\Theta_j$ and provides specific localized information.
This %framework 
means that, giving two area $i, j$ for two data distributions $\Theta_i$ and $\Theta_j$, 
 a sample $d'$ from $\Theta_i$ is %markedly
 distinctively dissimilar from a sample $d''$ from $\Theta_j$. 
 This dissimilarity can be quantified using a specific distance $m(d', d'')$, which %assesses 
 determines the disparity between two distributions.
Formally, giving an error bound $\delta$, the dissimilarity intra-area and inter-area can be quantified as:
\begin{equation}
    \forall i\neq j, \forall d,d' \in \Theta_i, \forall d'' \in \Theta_j: m(d, d') \leq \delta < m(d, d'')
\end{equation}

In $A$, a set of \emph{sensor nodes} $S = \{s_1, s_2, ... s_{{n}}\}$ ($n \gg |A|$) are deployed,
 each capable of processing data and participating FL.
 Sensors are assumed to be randomly uniformly distributed, and we let the position of a sensor $s_i$ be $p_i = (x_i, y_i)$: 
 each sensor will be located in a specific area $a_j$, though this information is not available to sensors.
 Moreover, each node has a certain communication range $r_c$ and hence can communicate only with nodes within that range:
 this form a neighbourhood $N_i$ for each node $i$.
%
% Initially, all sensors share a common neural network model $nn$ which is trained using the locally perceived data $d_j$.
%
Locally, 
each node $i$ creates a dataset $D_i$ from samples perceived from the data distribution $\Theta_j$. 
In this work, we consider a general classification task where each sample $d_i$ in the data distribution $\Theta_j$ consists of a feature vector $x_i$ and a label $y_i$. 
Therefore, the complete local dataset $D_i$ is represented as $D_i = \{(x_1, y_1), (x_2, y_2), \ldots, (x_m, y_m)\}$.
This dataset is used to train a local model $nn_i^t$ where $t$ is the time step of the learning process.
The $nn_i^0$ is initialized with a common neural network model $nn$ shared among all nodes.
%
 %As $\Theta_j$ is unique to a specific area $a_j$, the nodes can utilize this information to discern areas and nodes transmitting neural networks from there in federation steps. %This is achieved by the dissimilarity metric $m_{i,j}$.
%The sensor nodes, however, do not know in which area they are located, and they need to infer this information based on the data they perceive. 
%
For sake of simplicity, we limited the learning process in $T$ time steps, 
 where at each time step $t$, each node $s_i$:
 \begin{itemize}
    \item perform local training on its local dataset $D_i$ to get a local model $nn_i^t$;
    \item can share its model to one (or all) of its neighbours in $N_i$
 \end{itemize}
For each time step $t$, the nodes should devise a set of \emph{federations} $F^t = \{f_1^t, f_2^t, \ldots, f_m^t\}$, where each $f_j$ is characterized by a leader node $l_j$ that is responsible for the federation
 and a group of nodes that are part of the federation.
Giving $T$ the number of time steps, the goal of \approach{} is to create a set $F^T$ which approximates the areas $A$ and finds a federation-wise model $nn_j^T$ for each federation $f_j^T$ (i.e., the one created by the leaders). 
Let $nn^T = \{nn_1^T, nn_2^T, \ldots, nn_m^T\}$ be the set of all federation-wise models.
% that minimizes the error within each area and across all available areas.
 %Conceptually, if the nodes are infinitely dense, each $f_j$ would occupy a single area $a_j$.
 %where each federation $f_j$ is composed of nodes that are located in the same area $a_j$. 
 %Each federation has also a leader node $l_j$ that is responsible for the federation itself 
 %and a group of nodes that are part of the federation.
%Each federation $f_j$ within an area creates a shared federation-wide model $nn_j^T$,
% with the goal of minimizing the error within each area and across all available areas. %area wise.

%The goal of this approach is to create federations that are coherent with the local data distribution (i.e., minimizing the variance in the data distribution within federations)
% and to minimize the error within each area and across all available areas.
%

%The former can be formally described as:
%\begin{equation}
%\min_{f_j} Var(\{D_i | i \in f_j\})
%\end{equation}
%This goal ensures that each federation \( f_j \) is characterized by low variance in the data %distributions \( \Theta_i \) among its nodes, 
 %which helps in achieving more consistent and reliable model training and performance across similarly situated nodes.
Given a correct federations partitioning, we want to find a set of federation-wise models that minimize the error across all areas, 
 which can be formally described as:
\begin{equation}
    \min_{nn^T} \sum_{nn_i^T \in nn^T} L(\Theta_i, nn_i^T)
\end{equation}
Where $L$ represent the average error computed by a loss function from each sample $x_i$ and label $y_i$ in the dataset $\Theta_i$ using the model $nn_i^T$.
%where $L$ represents a loss function indicating how well $nn_i^T$ represents $\Theta_i$ in the given classification task, namely how well the model is able to predict the labels $y_i$ given the features $x_i$ sampled from ${\Theta_i}.  
%A low value of $L$ indicates a good match between $nn_j^T$ and $\Theta_i$.
%To achieve this, we have to minimize the number of nodes contributing to the shared $nn_i^T$ that are not subject to $\Theta_i$.

\subsection{Proposed solution}
\begin{figure*}
    \fontsize{8.0pt}{10pt} \selectfont
    \def\svgwidth{\linewidth}
    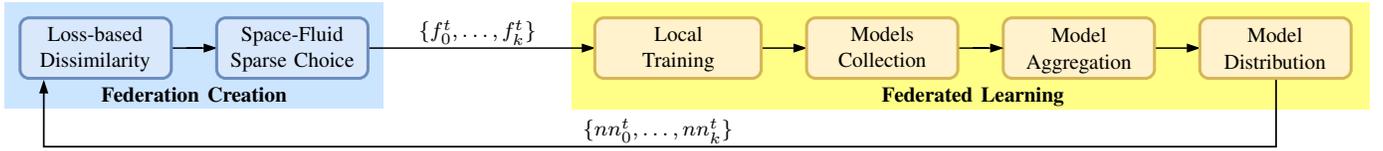
    \caption{Overview of the proposed algorithm. }\label{fig:algorithm-overview}
\end{figure*}
% \begin{figure}
%     \centerline{\includegraphics[width=0.5\textwidth]{figures/zones.pdf}}
%     \caption{ Graphical representation of the problem.
%     In this case, there is an area $A$ divided into four sub-areas $\{a_1, a_2, a_3, a_4\}$. 
%     Each sub-area has a different data distribution (represented by the colors). 
%     The white nodes represent sensors $S = \{s_1, ..., s_n\}$, while links between sensors represent
%     the neighborhood relation (i.e., the ability of two sensors $s_i$ and $s_j$ to exchange information). 
%     }
%     \label{fig:zones}
% \end{figure}

Our goal is to effectively organize the sensor nodes into federations that enhance learning efficacy by exploiting federation-wise data similarities. 
 To achieve this, we utilize a distributed federated learning approach structured with SCR~\cite{DBLP:conf/coordination/CasadeiPVN19} (see \Cref{sec:scr}) in which the regions are the federations of the learning process.
 %This pattern facilitates efficient regional data handling and decision-making among nodes within the same federation.  
%We structure our approach (summarized in \Cref{fig:algorithm-overview}) into two distinct macro phases:
%\begin{enumerate}
%    \item \textbf{Federation Creation:} nodes determine their federation memberships based on local data distributions.
%    \item \textbf{Federated Learning:} nodes collaboratively train models by exchanging and updating based on their neighbors' models.
%\end{enumerate}

%\gianluca{We should add the AC code here?}
%\paragraph{Federated Learning}
%The federated learning framework is structured around a well-established self-organizing system pattern known as the \emph{Self-organizing Coordination Region}~\cite{DBLP:conf/coordination/CasadeiPVN19}. 
% This pattern facilitates efficient regional data handling and decision-making among nodes within the same federation. 
The \approach{} process can be roughly described in four steps: %is delineated as follows:

\begin{enumerate}
    \item \emph{Federation Creation:} \label{stp:fedCreate} Nodes form a federation via a distributed multi-leader election process. %(see next section). 
    The elected leaders are responsible for the coordination and management of federation activities.
    \item \emph{Models Collection:} The leaders collect models from each node within the federation. 
    \item \emph{Model Aggregation:} Based on the aggregated models, 
    the leaders synthesize a unified federation-wise model that reflects the collective learning outcomes of the federation.
    \item \emph{Model Distribution:} The synthesized model is then disseminated by the leaders to all nodes in the federation, 
    ensuring that each node operates with the most updated and accurate model reflective of their shared data environment.
\end{enumerate}
This is further detailed in \Cref{fig:algorithm-overview}.
%\gianluca{better describe when the model is updated with the respect to the federated one.}

%\paragraph{Federation Creation}
To autonomously establish federations (Step \ref{stp:fedCreate} of the \ac{FL} process) aligning with local data distributions, 
 we utilize a decentralized, multi-leader approach inspired by the ``space-fluid'' sparse choice algorithm \cite{DBLP:journals/lmcs/CasadeiMPVZ23}. 
 In this method, federations form around a nominated leader,
  based on the leader's proximity to the nodes and the dissimilarity of their data distributions.
Indeed, federations expand and contract dynamically based on the accumulated \emph{dissimilarity} of their data distributions, computed from the nodes' models. 
 Specifically, we consider a loss-based dissimilarity measure, 
 denoted as \(ds(i, j)\), calculated by \(L_{i, j} + L_{j, i}\). 
 Here, \(L_{i, j}\) and \(L_{j, i}\) represent the average loss function using local dataset \(D_i\) and the model from the adjacent node \(nn_j\), respectively.
Formally:
\begin{equation}
    L_{i, j} = L(D_i, nn_j) = \frac{1}{|D_i|} \cdot \sum_{(x, y) \in D_i} \ell(f(x; nn_j), y)
\end{equation}
\(f(x; nn_j)\) is the prediction of the model \(nn_j\) on the input \(x\), and \(\ell\) one of the standard loss functions (e.g., cross-entropy, mean squared error).

We introduce a gradient field \(G: N \rightarrow \mathbb{R}\)~\cite{DBLP:conf/saso/AudritoCDV17},
 constructed based on the dissimilarity metric \(ds(i, j)\) 
 and using the leaders of the federations as source zones. 
This constructs a potential field where, for each node, 
 the shortest accumulated error to the leader is defined. 
For example, consider a simple ring topology composed of three nodes, \texttt{a} - \texttt{b} - \texttt{c}, where \texttt{a} is the leader of the federation.
The gradient field \(G(c)\) is then computed as the sum of the errors from \texttt{c} to \texttt{b} and from \texttt{b} to \texttt{a}, namely: \(G(c) = ds(c, b) + ds(b, a)\).
 Thus, a node can decide to join a federation based on the error relative to the leader, 
 effectively creating a federation bounded by a maximum path error \(\sigma\).
In this approach, this gradient field is continuously updated, 
 ensuring that the federations align closely with data distributions.

Starting from this notion of accumulated error, we can better illustrate the federation creation process, consider the following steps:
\begin{enumerate}
    \item Each node proclaims its candidacy for becoming a leader within its local neighbourhood.
    \item Nodes propagate the leadership candidacy information to their neighbours, 
    including the candidate's model and the error metric computed from the leader to the candidates (i.e., computing $G$).
    \item Nodes evaluate received candidacies and disregard those where the path error from the leader exceeds a predefined threshold $\sigma$. %(look at \Cref{fig:federation-creationg}). 
    %This threshold is determined based on domain-specific knowledge and represents the maximum tolerable error.
    \item In scenarios where multiple valid candidacies are recognized, 
    nodes employ a \emph{competition policy} to select the most suitable leader, 
    ensuring optimal federation configuration based on local data characteristics.
\end{enumerate}
The competition policy can be based on various criteria, such as the quality of the model, the error metric, or just the node identifier. 
In this solution, we both use the accumulated error metric and the node identifier to break ties.
Leveraging this space-fluid sparse choice approach, the following statement holds: for every node \(i\) in federation \(f_j\), the condition \(G(n_i) \leq \sigma\) is satisfied. 
By choosing an appropriate \(\sigma\), we ensure that the federations are coherent with the perceived data distributions. Indeed, even if it is not possible to guarantee a direct relation between \(\sigma\) and \(\delta\), the latter can be used as a reference to determine the former: higher \(\delta\) values require higher \(\sigma\) values to ensure that a federation $j$ aligns with the data distribution $\Theta_i$ of the underlying area $i$.
%\begin{figure}
%    \centering
    % \includegraphics[width=\linewidth]{overall-idea.drawio.pdf}
%    \fontsize{8.0pt}{10pt} \selectfont
%    \def\svgwidth{\linewidth}
%    \input{figures/overall-idea.pdf_tex}
%    \caption{The diagram depicts the decentralized federation formation where bold nodes
%    nominate themselves as leaders forming three federations (represented by different colours). 
%    Arrows inside the rectangles show information flow between nodes. 
%    Distances marked by $\sigma$ represent the maximum permissible error metrics for leader selection.
     %ensuring that federations align closely with local data distributions while maintaining data privacy.
%     }\label{fig:federation-creationg}
%\end{figure}

The application of this pattern is structured to promote federated learning and adaptation:
	\begin{itemize}
		\item \emph{Continuous Model Exchange:} At each time step, nodes exchange their models with neighbours. 
		This exchange includes the integration of newly received models into each node's local training processes, 
		thereby enhancing the model's accuracy and relevance.
		\item \emph{Specialized training:} 
        The process of repeatedly exchanging and updating the model creates a continuous loop that improves the model's training across a specific network. 
        This loop gradually enhances the accuracy and efficiency of the model, allowing it to better represent areas with similar data characteristics and to create smaller, more focused models compared to the overall global model. 
        Additionally, it increases the model's ability to make predictions in these areas by continuously incorporating new insights from similar data experiences.
	\end{itemize}
	
	By leveraging this structured approach to federated learning, 
	we ensure that each federation not only optimizes its internal operations but also contributes effectively to the overarching goal of achieving high-accuracy, 
	federation-specific models within the distributed learning network.
%The core of our federation creation strategy hinges on the use of a \emph{dissimilarity measure} \(ds_{i, j}\), 
% calculated without direct data sharing. 
% This measure assesses the difference between the local data distribution \(\Theta_i\) 
% and the model \(nn_j\) of a neighboring node \(j\). 
% Although direct measures like the Jensen-Shannon divergence~\cite{menendez1997jensen} or the Kullback-Leibler divergence~\cite{kullback1951information} could be employed if raw data were available, 
% our approach only requires the exchange of model outputs, preserving data privacy.

%\gianluca{perhaps we can drop that.}
%The dissimilarity measure \(m_{i, j}\) is defined by the properties:
%\begin{itemize}
%    \item \textbf{Zero-diagonal:} \(m_{i, i} = 0\)
%    \item \textbf{Non-negativity:} \(m_{i, j} \geq 0\)
%    \item \textbf{Symmetry:} \(m_{i, j} = m_{j, i}\)
%\end{itemize}

%Giving $L_{i, j} = L(D_i, nn_j)$ the loss function between the local data distribution $D_i$ and the model $nn_j$, 
%We implement a loss-based dissimilarity measure for \(ds_{i, j}\),
%  calculated as \(L_{i, j} + L_{j, i}\), where \(L_{i, j}\) and \(L_{j, i}\) represent the loss functions using local data \(D_i\) and the model from the adjacent node \(nn_j\), respectively. 

%This approach facilitates the federative decision-making process by enabling nodes to determine their alignment based on the similarity of model performance, 
%  thus fostering federations that are coherent in terms of data distribution characteristics without compromising data privacy.
%
\begin{algorithm}
    \caption{Proximity-based self-federated learning}
\label{alg:self_fed_learning}

\DontPrintSemicolon
%\SetAlgoLined
\SetKwInOut{Require}{Require}
\Require{Each node $s_i \in S$ has an initial model $nn_i^0$}

\For{each time step $t = 1$ to $T$}{
    \For{each node $s_i$ in $S$}{
        $nn_i^t \leftarrow$ LocalTraining($D_i$)\;
        Broadcast $nn_i^t$ to neighbours $N_i$\;
        Select federation $j$ based on the dissimilarity metric $ds$ accumulated from neighbours (using a \(G\) field)\;
        \If{$s_i$ is a leader}{
            Collect models (using \(C\)) from federation members and aggreate them to a new model $nn_{f_j}^t$\;
            Distribute $nn_{f_j}^t$ to each $s \in f_j$ (using \(G\));
        }
        Receive models from federation leader and them updating the local model $nn_i^{t+1}$\;
    }
}

\end{algorithm}

% \noindent\begin{minipage}{\textwidth}
%     \begin{lstlisting}[language=scafi]
    
%     rep(init())((local, global) => { 
%         val metric = lossBasedMetric(global)
%         val aggregators = S(threshold, metric) // Dynamic aggregator selection
%         local.evolve() // 1. Local training step
%         val pot = classicGradient(aggregators, metric) // Potential field for model sharing
%         val sender = G(potential, metric, mid(), (_) => nbr(mid()))
%         // 2. Model sharing 
%         val info = CWithSenderField[Set[Model]](_ ++ _, Set(local), Set.empty, sender) 
%         // 3. Model aggregation
%         val aggregateModel = aggregation(info) 
%         // 4. Global model update
%         sharedModel = broadcast(aggregators, aggregateModel, metric) 
%         mux(impulsesEvery(epochs))
%         {(combineLocal(evolvedModel, sharedModel), combineLocal(evolvedModel, sharedModel))} 
%         {(evolvedModel, combineLocal(evolvedModel, sharedModel))}
%     })
% \end{lstlisting}
% \end{minipage}

%\meta{section length 2 pages}
\section{Experimental Evaluation}\label{sec:evaluation}
\begin{figure*}
    \centering
    \begin{subfigure}[b]{0.24\textwidth}
        \includegraphics[width=\linewidth]{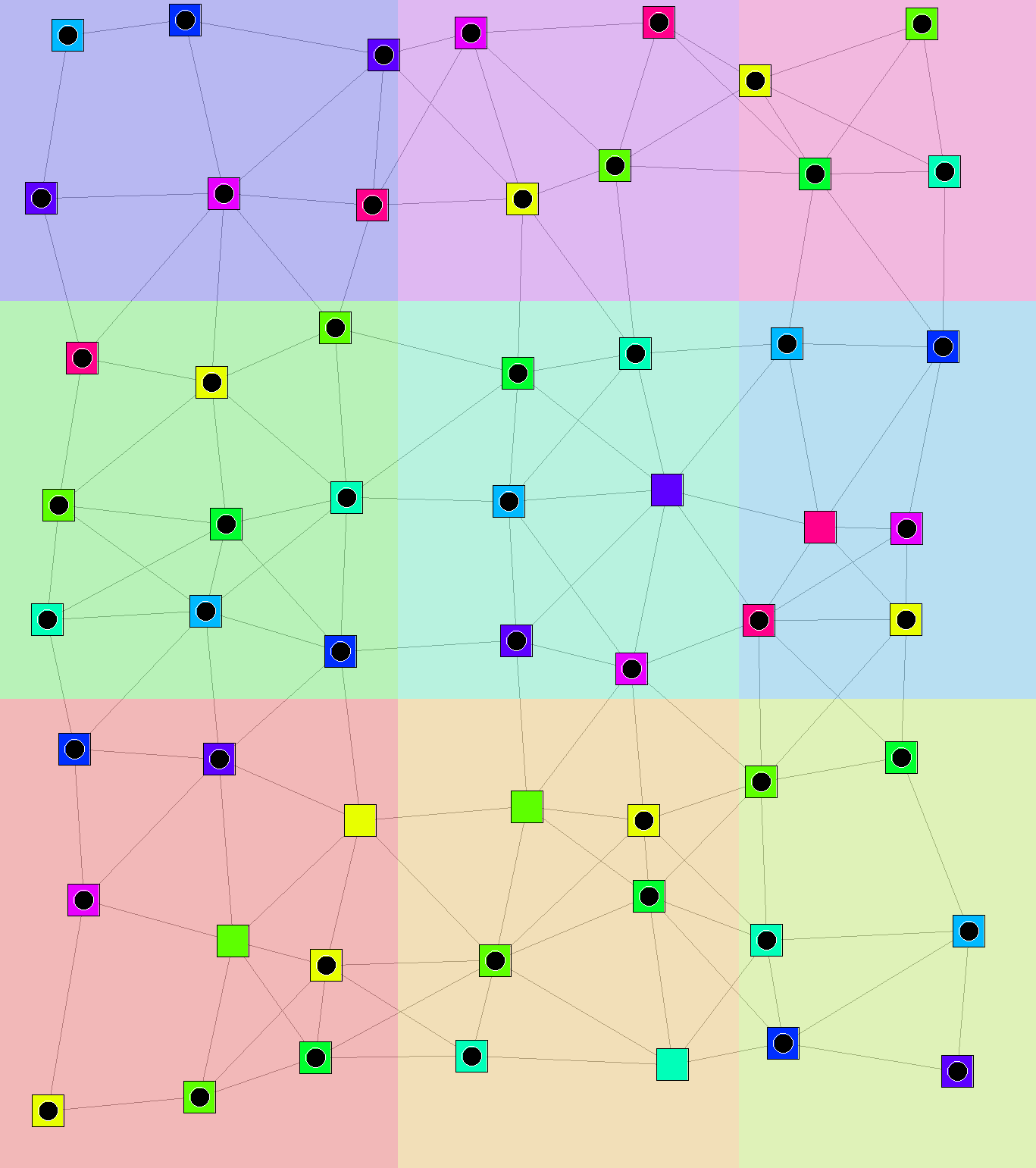}
    \end{subfigure}
    \begin{subfigure}[b]{0.24\textwidth}
        \includegraphics[width=\linewidth]{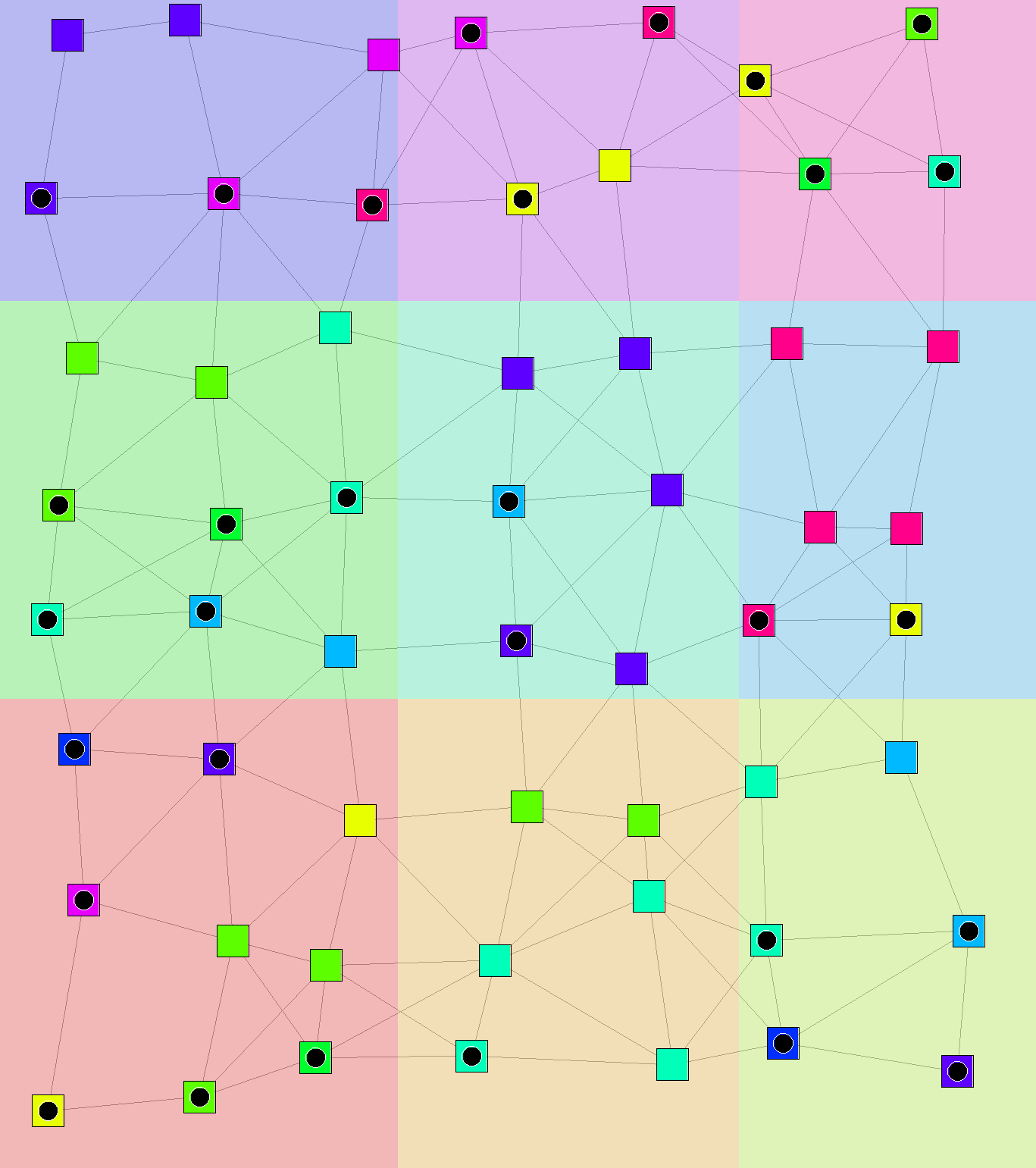}
    \end{subfigure}
    \begin{subfigure}[b]{0.24\textwidth}
        \includegraphics[width=\linewidth]{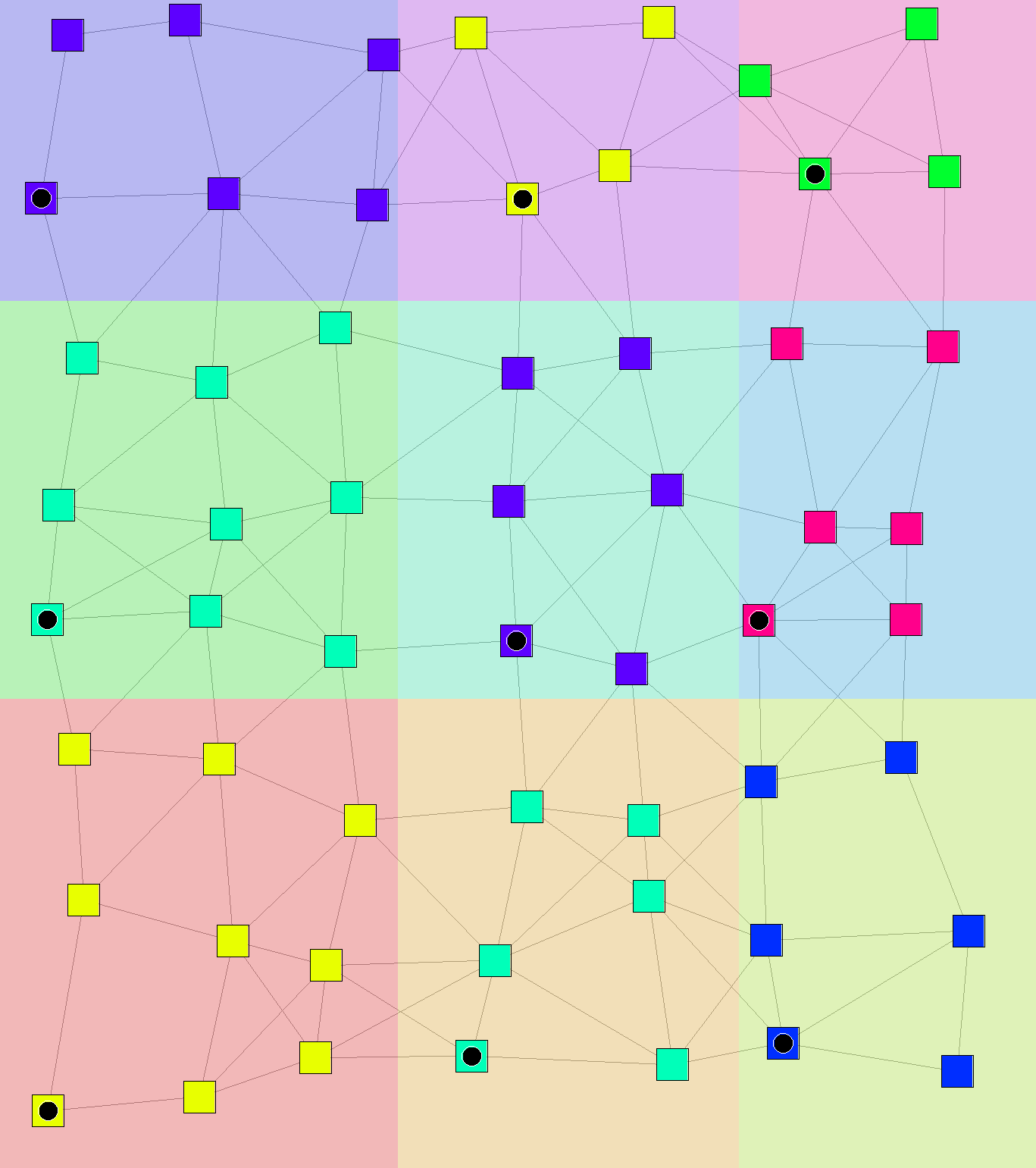}
    \end{subfigure}
    \caption{A simulation run of \approach{} with $|A| = 9$. 
    The background colour represents the different areas, while the colour of the nodes represents the federation they belong to.
    Leaders are represented with an inner black circle.
    In the first snapshot, all the nodes elect themselves as leaders, then the nodes start to form federations based on the dissimilarity metric and finally the federations are stable and aligned with the areas.
    }\label{fig:simulation-snapshots}
\end{figure*}

\subsection{Learning setup}
To evaluate the effectiveness of our approach, we performed learning on a well-known dataset in computer vision, namely: Extended MNIST~\cite{DBLP:conf/ijcnn/CohenATS17}.
    Particularly, we used the version with handwritten letters, which contains $124800$ train samples and $20800$ test samples from $26$ classes.
    Data have been synthetically partitioned to obtain multiple non-IID distributions and simulate labels skewness (i.e., a device has only a subset of the labels). 
    More specifically, we conducted several simulations with a variable number of areas 
    $|A| \in \{3, 5, 9\}$. 
    %In this way, a device located in a certain area $a_i$ will perceive data with a distribution different from another device situated in an area $a_j$. 
    We performed a synthetic partitioning of the data -- as done in other works in the literature such as~\cite{DBLP:journals/corr/abs-2302-02949,DBLP:conf/icde/LiDCH22} -- 
    rather than using real-world non-IID datasets, 
    to be able to control the degree of label skewness and to simulate various situations with different numbers of areas, enabling a more extensive evaluation.

The learning process was conducted using both the algorithm proposed by us and the classic centralized FedAVG (utilized as a baseline for comparisons). 
    The same hyperparameters were employed for both approaches. 
    A simple MLP neural network with $128$ neurons in the hidden layer was trained for a total
    of $60$ global rounds. 
    At each global round, every device performed $2$ epochs of local learning using a batch size of $64$, 
    the ADAM optimizer with a learning rate of $0.001$, and a weight decay of $0.0001$. 
    Additionally, in the \approach{} algorithm proposed, 
    FedAVG was used to aggregate the models within the various federations.

All simulations\footnote{
        \url{https://github.com/nicolasfara/experiments-2024-ACSOS-opportunistic-federated-learning}.
        %note for the reviewers: we anonymized the repository with the code to preserve the double-blind review process.
    } were conducted using the Alchemist simulator~\cite{Pianini_2013}, which facilitated the division of data across various areas, 
    spatial distribution of devices, and their neighbourhood relationships. 
    The \approach{} was implemented using ScaFi~\cite{DBLP:journals/softx/CasadeiVAP22}---a macroprogramming language for aggregate computing combined with PyTorch for the neural network training.
    For the baseline algorithm, we conducted $20$ simulations with varying seeds. Conversely, for the \approach{}, we conducted $180$ simulations varying: 
    i) the seed (from $0$ to $19$); 
    ii) the number of areas ($3$, $5$, and $9$); and 
    iii) the loss threshold $\sigma$ ($20$, $40$, and $80$) used to determine the maximum expansion of a federation.

Moreover, we conducted $10$ additional simulations with nine areas in which we explored the effects of introducing a new node into an already stable system. 
 The primary goal was to evaluate the system's ability to self-adapt without disrupting the existing federations. 
A special node, denoted as $\bigstar$, was introduced to traverse all nine areas in sequence. 
 The movement pattern of $\bigstar$ includes systematically moving from one area to the next, covering the entire set of areas. 
 Upon entering each new area, $\bigstar$ remains there for a fixed period of 5 time steps to simulate interaction and impact on the area.
The total duration of the simulation is 140 time steps.
\subsection{Metrics}\label{sec:metrics}
Regarding the learning process, we have extrapolated the loss and the accuracy for each node. 
The loss used in this case is a negative log-likelihood loss which is defined as:
\begin{equation}
NLL(D, nn) = -\sum_{i=1}^N y_i \log(nn(x_i)), \quad \text{where } x_i, y_i \in D
\end{equation} 
where \( nn(x_i) \) is the predicted probability of the true label \( y_i \).
The accuracy was calculated as:
\begin{equation}
\text{Accuracy} = \frac{\text{Number of correct predictions}}{\text{Total number of predictions}}
\end{equation}
These values have been evaluated for both the training set and the validation set. 
During the testing phase, only the accuracy was verified since it is the metric of interest in our classification problem.

Regarding the federations, we have evaluated the number of federations created and their correctness.
The federation count $|F|$, can be described as the unique number of leaders in the system at a given time step.
Formally, given a set of leaders $L = \{l_1, l_2, \ldots, l_m\}$, the federation count is $|F| = |L|$.
The correctness of federations is assessed based on the location concurrence of nodes within a single federation. 
Mathematically, the correctness metric for a federation $f_j$ is expressed as:
\begin{equation}
\Diamond(f_j) = \sum_{n \in f_j} \chi(\xi(n), \xi(l_j))
\end{equation}
where $\xi(n)$ denotes the actual area of node $n$, and $\xi(l_j)$ is the area of the federation leader $l_j$. 
The function $\chi$ outputs $0$ when both areas are the same, 
indicating correct placement, and $1$ when they differ.
Therefore, $\Diamond(f_j)$ calculates the count of nodes that are mistakenly part of federation $f_j$. 
A value of $0$ for $\Diamond(f_j)$ confirms that all nodes are correctly aligned, 
while any value greater than $0$ signals misalignments within the federation.

Finally, in our experiment involving node movement, 
we evaluated changes in leadership that a moving node \(n_i\) perceives to understand the stability of the federations.
At a given time \(t\), this metric evaluates whether the leader of node \(n_i\) has changed compared to its leader at time \(t-1\). Formally, this is defined as:
\begin{equation}
    DL(n_i, t) = \chi(\zeta(n_i, t), \zeta(n_i, t-1))
\end{equation}
where \(\zeta(n_i, t)\) identifies the leader of node \(n_i\) at time \(t\).
Therefore, this metric will result in \(1\) if the leader has changed and \(0\) otherwise.
%where $\mathbf{1}_{\{s_i \in a_j\}}$ is the indicator function that is $1$ if $s_i$ is in area $a_j$ and $0$ otherwise.
\subsection{Results}
\begin{figure*}
    \centering
    %%%%%%%%%%%%%%%%%%%%%%% Train loss %%%%%%%%%%%%%%%%%%%%%%%%%%%%%%%%%%%%%%
    \begin{subfigure}[b]{0.32\textwidth}
        \includegraphics[width=\linewidth]{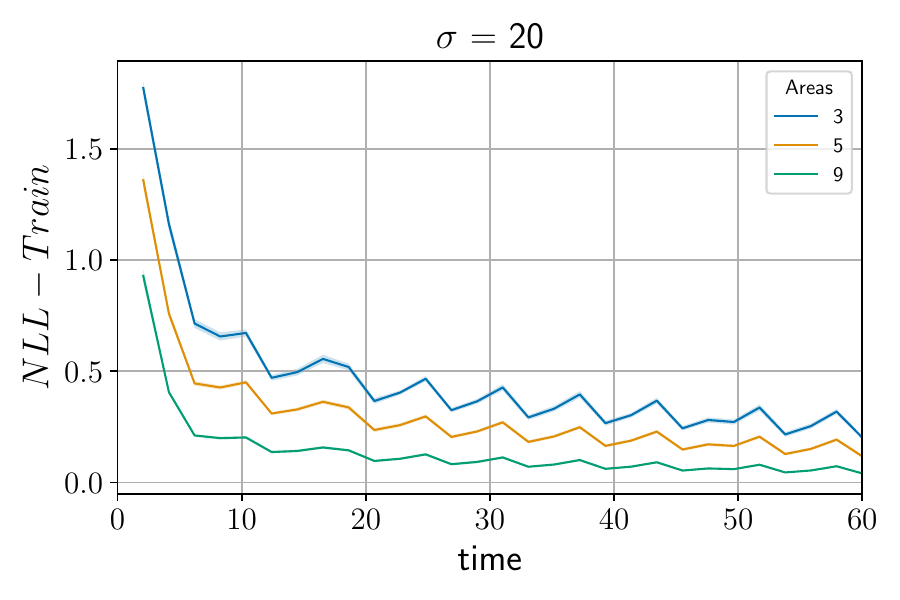}
    \end{subfigure}
    \begin{subfigure}[b]{0.32\textwidth}
        \includegraphics[width=\linewidth]{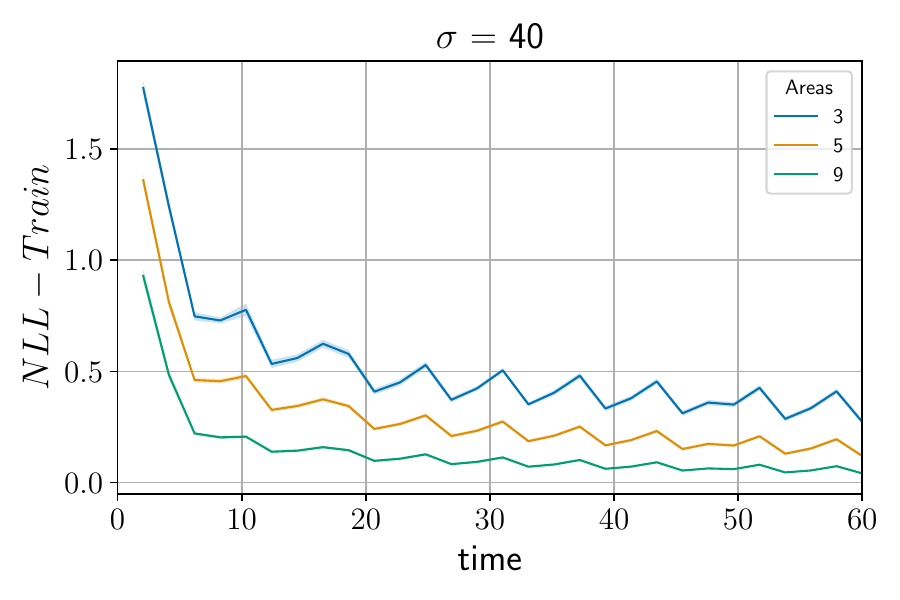}
    \end{subfigure}
    \begin{subfigure}[b]{0.32\textwidth}
        \includegraphics[width=\linewidth]{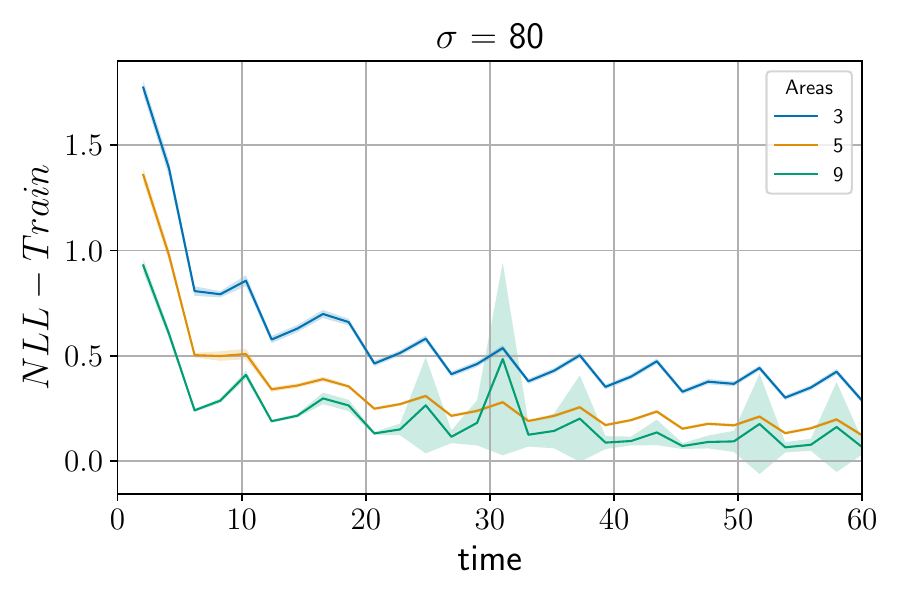}
    \end{subfigure}
    %%%%%%%%%%%%%%%%%%%%%% Validation loss %%%%%%%%%%%%%%%%%%%%%%%%%%%%%%%%%%%%%%%
    \begin{subfigure}[b]{0.32\textwidth}
        \includegraphics[width=\linewidth]{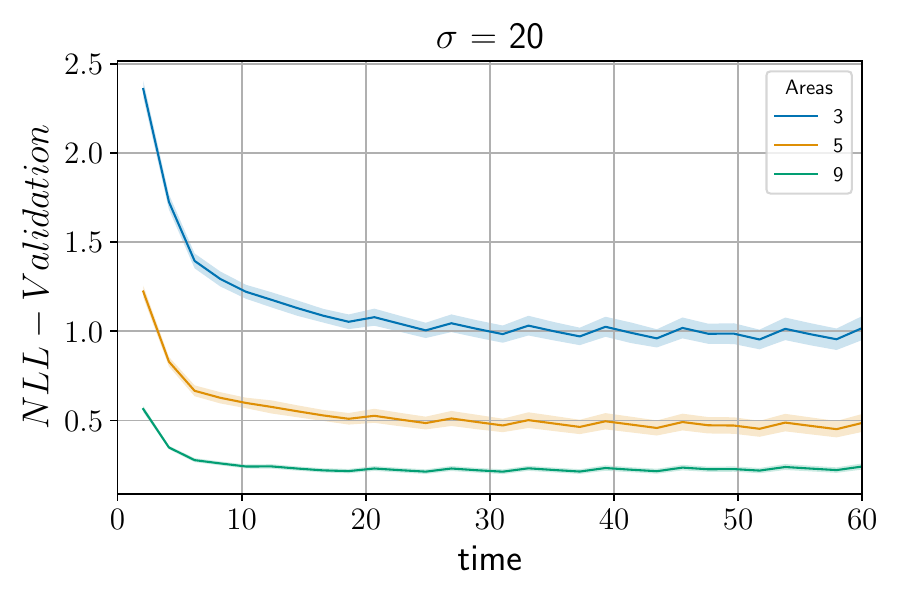}
    \end{subfigure}
    \begin{subfigure}[b]{0.32\textwidth}
        \includegraphics[width=\linewidth]{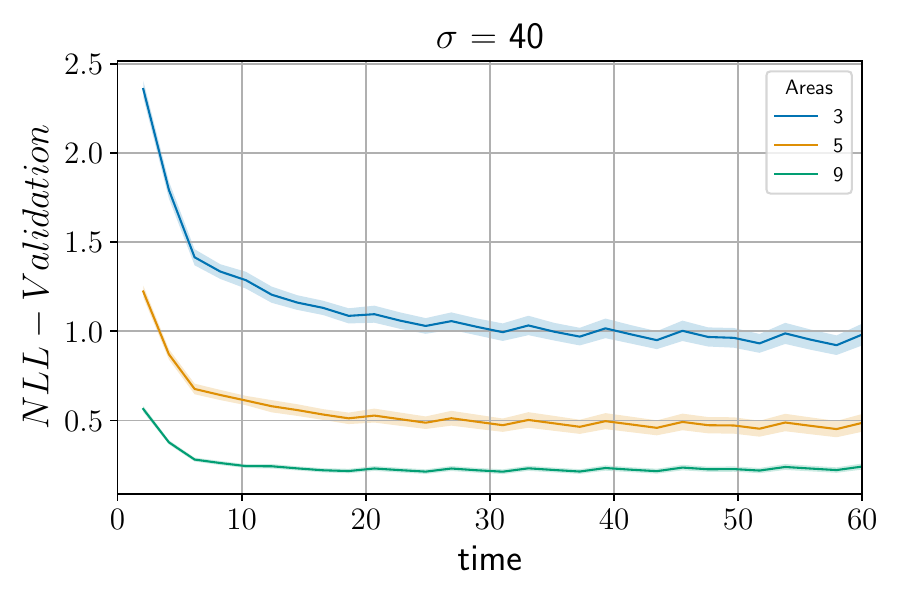}
    \end{subfigure}
    \begin{subfigure}[b]{0.32\textwidth}
        \includegraphics[width=\linewidth]{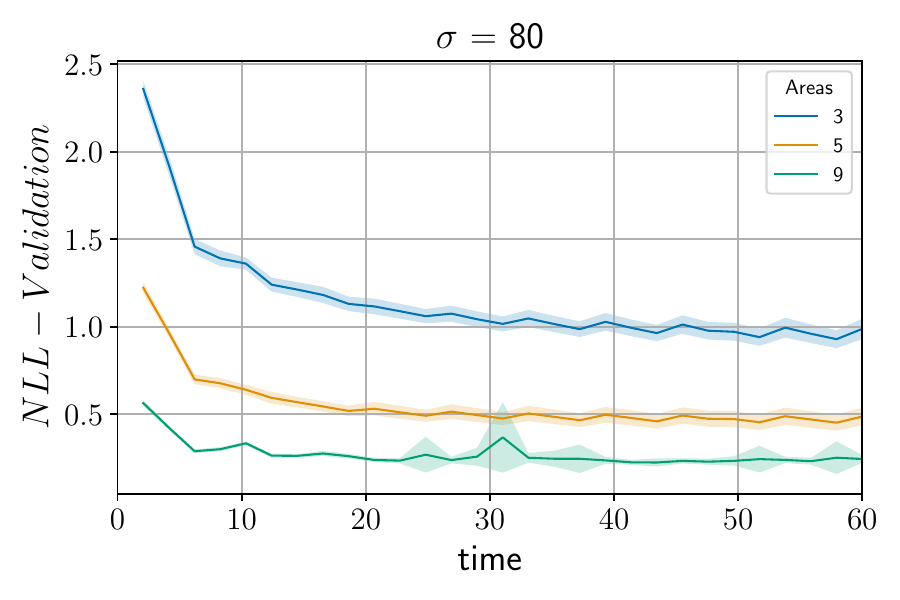}
    \end{subfigure}
    %%%%%%%%%%%%%%%%%%%%%% Validation Accuracy %%%%%%%%%%%%%%%%%%%%%%%%%%%%%%%%%%%%%%
    \begin{subfigure}[b]{0.32\textwidth}
        \includegraphics[width=\linewidth]{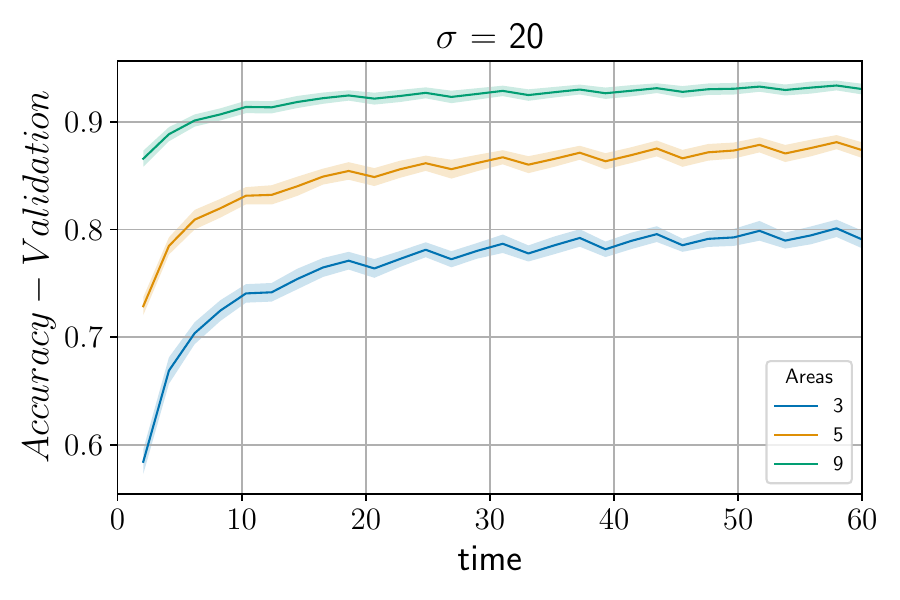}
    \end{subfigure}
    \begin{subfigure}[b]{0.32\textwidth}
        \includegraphics[width=\linewidth]{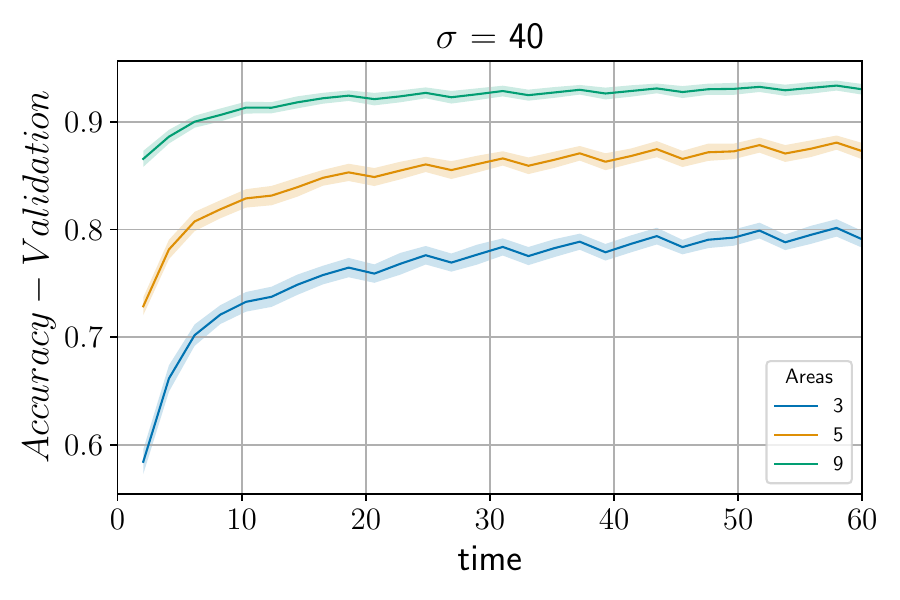}
    \end{subfigure}
    \begin{subfigure}[b]{0.32\textwidth}
        \includegraphics[width=\linewidth]{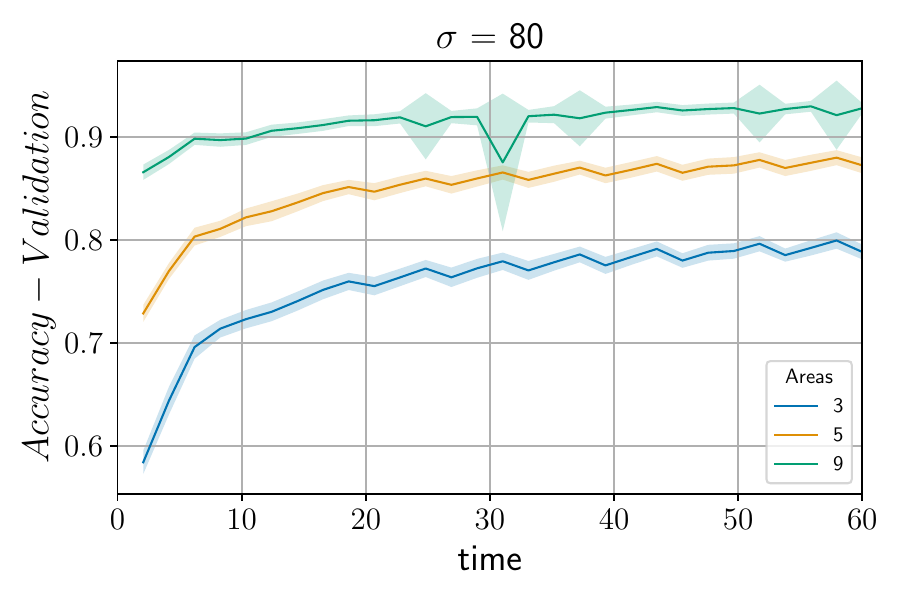}
    \end{subfigure}
    %%%%%%%%%%%%%%%%%%%%%%% Area Count %%%%%%%%%%%%%%%%%%%%%%%%%%%%%%%%%%%%%%
    \begin{subfigure}[b]{0.32\textwidth}
        \includegraphics[width=\linewidth]{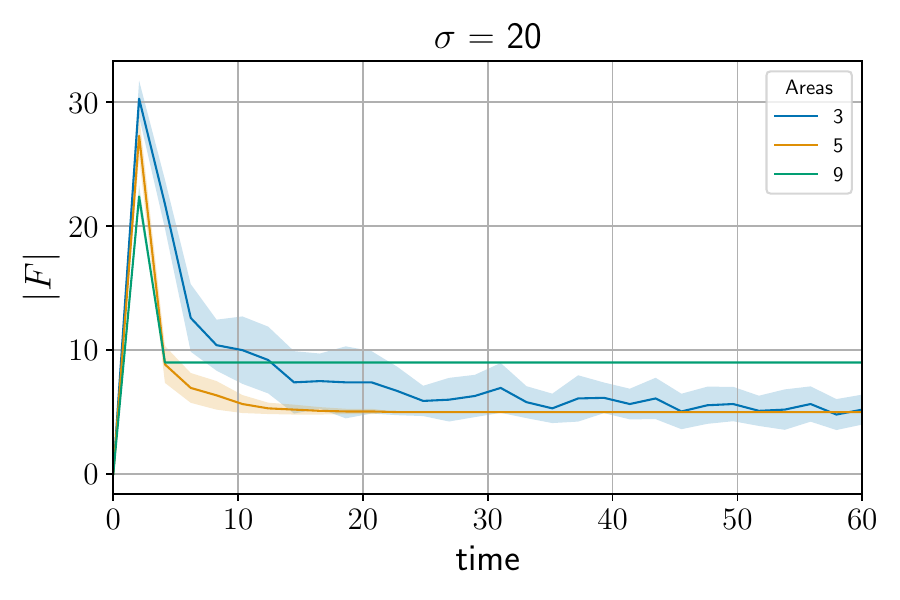}
    \end{subfigure}
    \begin{subfigure}[b]{0.32\textwidth}
        \includegraphics[width=\linewidth]{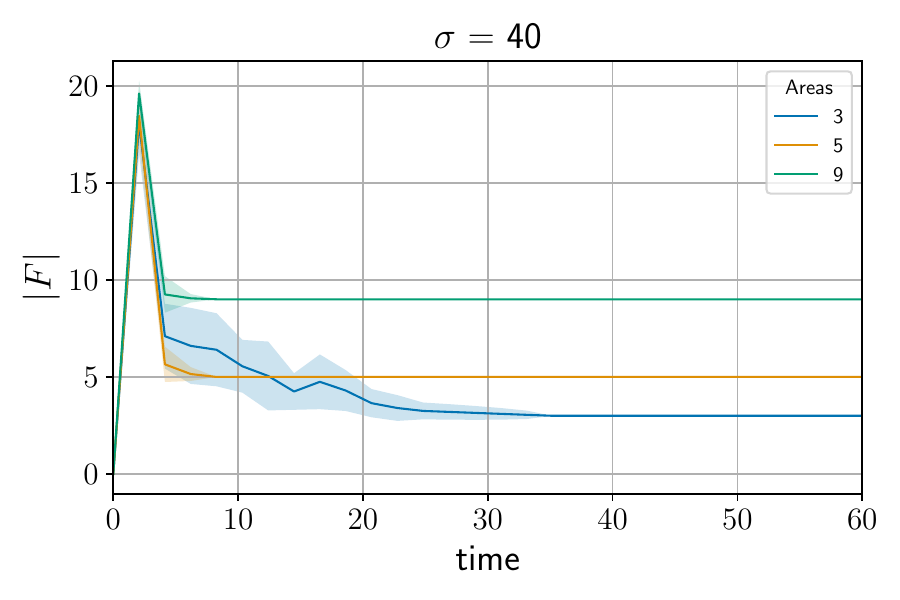}
    \end{subfigure}
    \begin{subfigure}[b]{0.32\textwidth}
        \includegraphics[width=\linewidth]{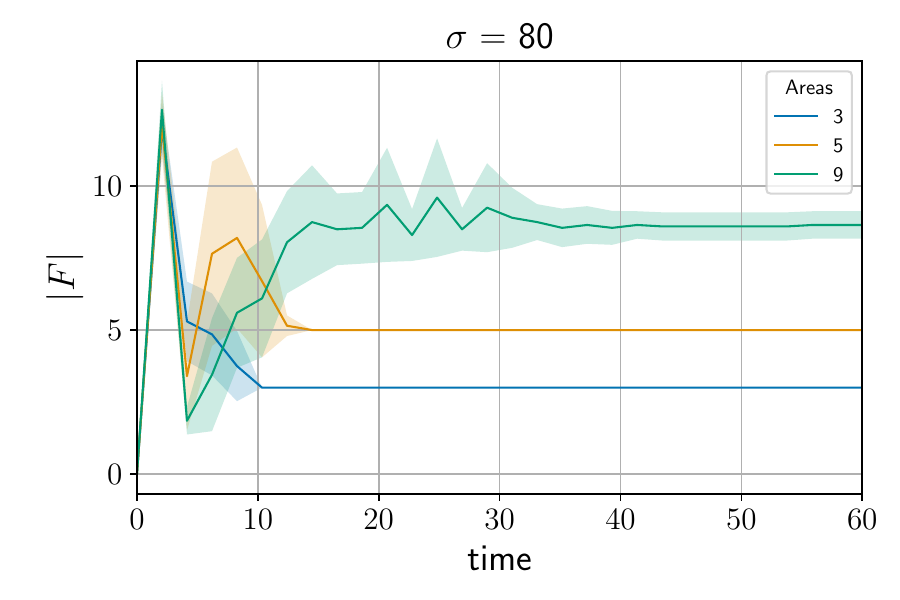}
    \end{subfigure}
    %%%%%%%%%%%%%%%%%%%%%% Area Correctness %%%%%%%%%%%%%%%%%%%%%%%%%%%%%%%%%%%%%%%
    \begin{subfigure}[b]{0.32\textwidth}
        \includegraphics[width=\linewidth]{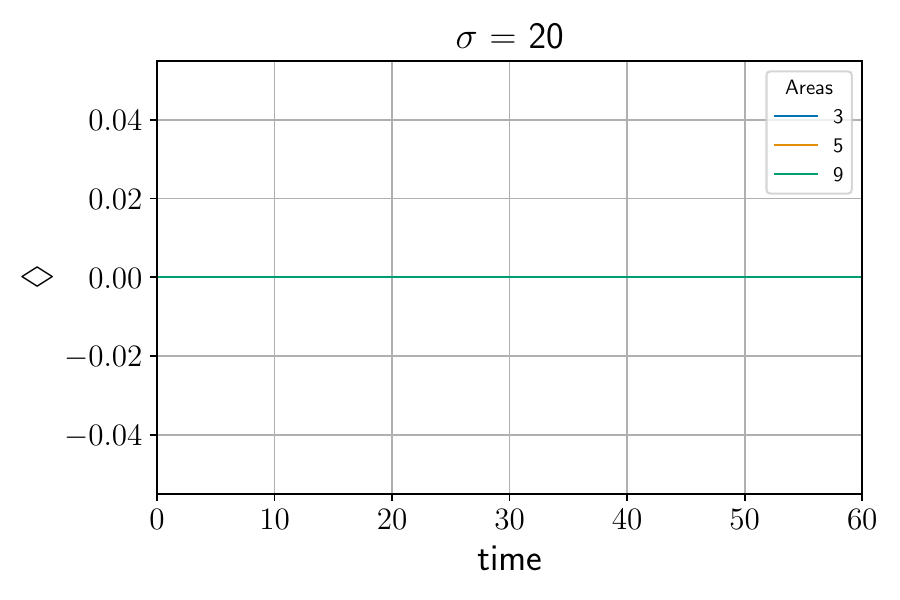}
    \end{subfigure}
    \begin{subfigure}[b]{0.32\textwidth}
        \includegraphics[width=\linewidth]{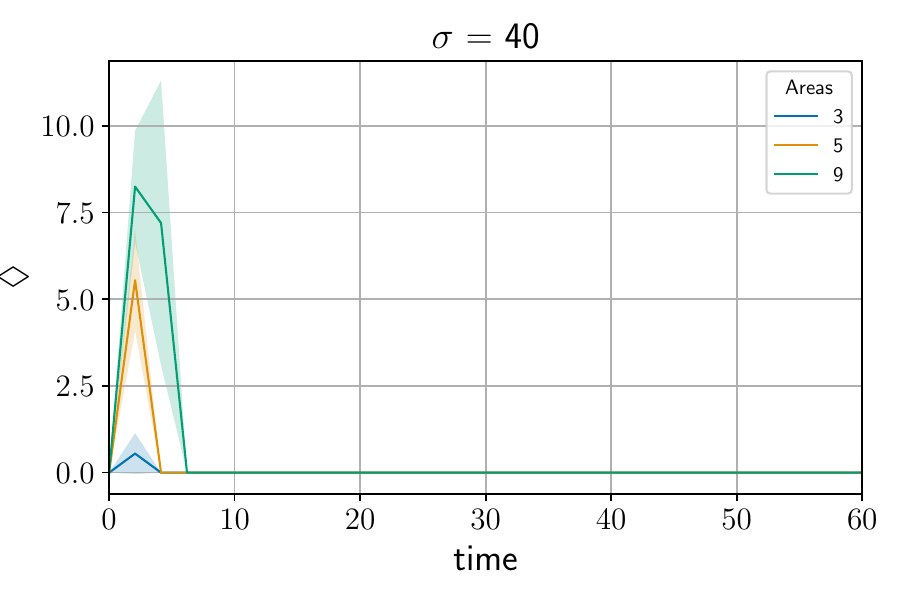}
    \end{subfigure}
    \begin{subfigure}[b]{0.32\textwidth}
        \includegraphics[width=\linewidth]{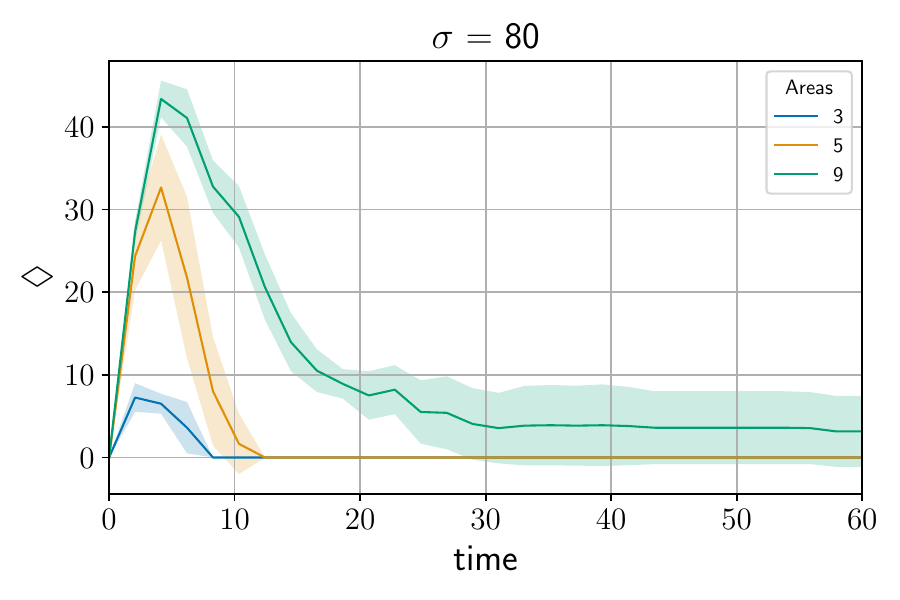}
    \end{subfigure}
    \caption{
        Data collected during training and validation. 
        Each column represents a different loss threshold, while rows represent metrics explained in~\Cref{sec:metrics}. We can see that the number of areas and the loss threshold have a significant impact on the performance of the system. \approach{} better performs with a higher number of areas and a lower loss threshold.
        }
        \label{fig:training-performance-study}
\end{figure*}

\begin{figure*}
    \centering
    \begin{subfigure}[b]{0.32\textwidth}
        \includegraphics[width=\linewidth]{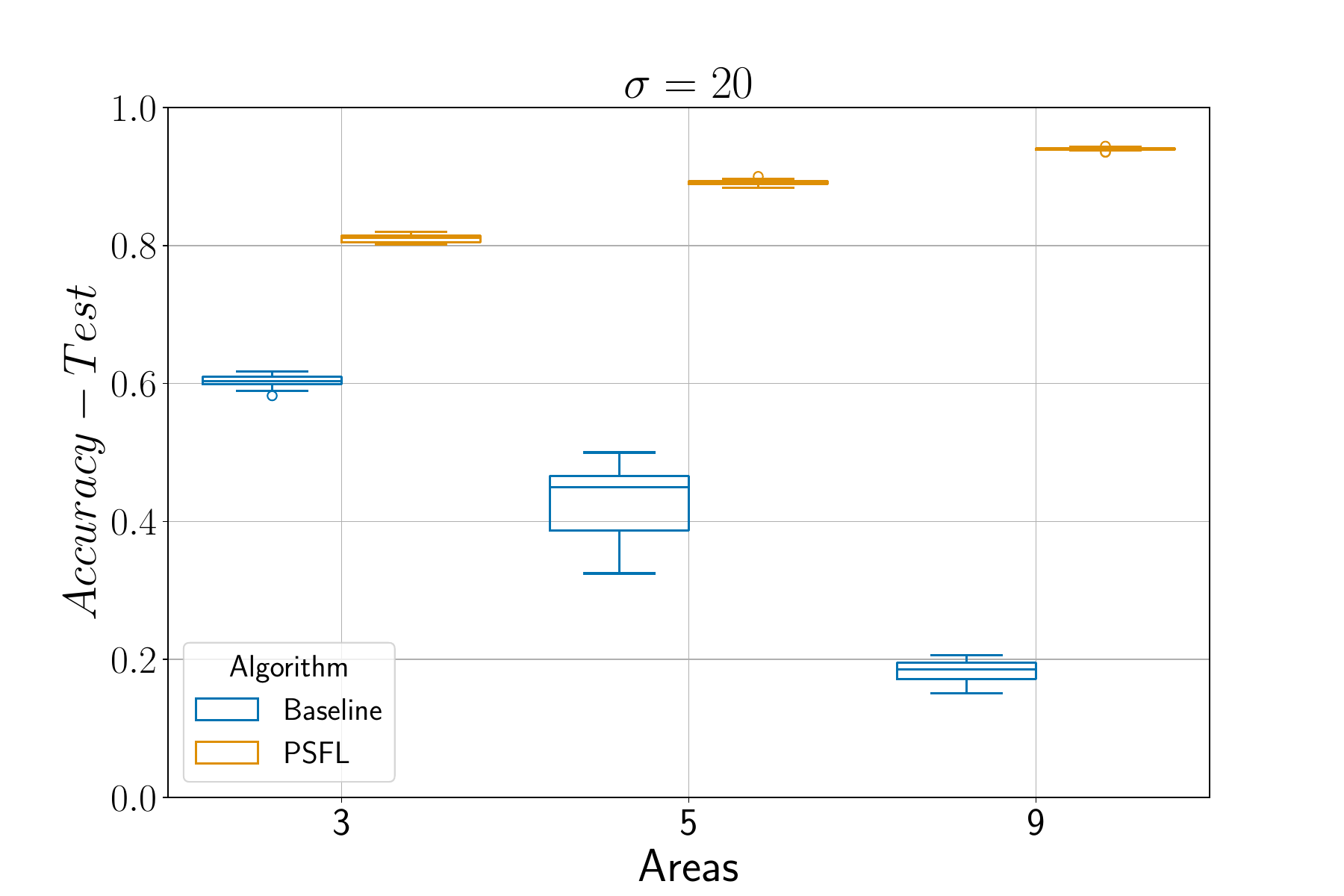}
    \end{subfigure}
    \begin{subfigure}[b]{0.32\textwidth}
        \includegraphics[width=\linewidth]{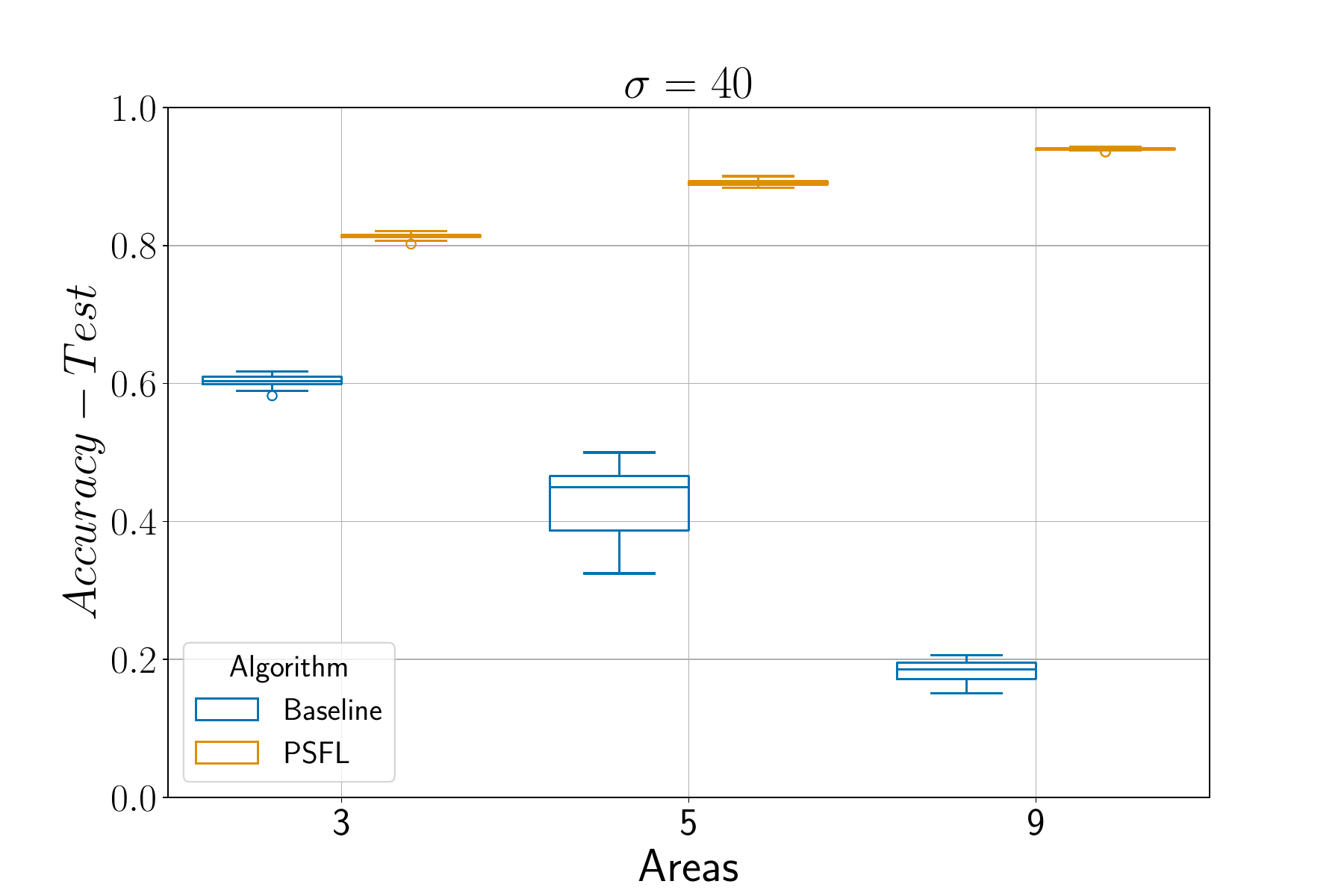}
    \end{subfigure}
    \begin{subfigure}[b]{0.32\textwidth}
        \includegraphics[width=\linewidth]{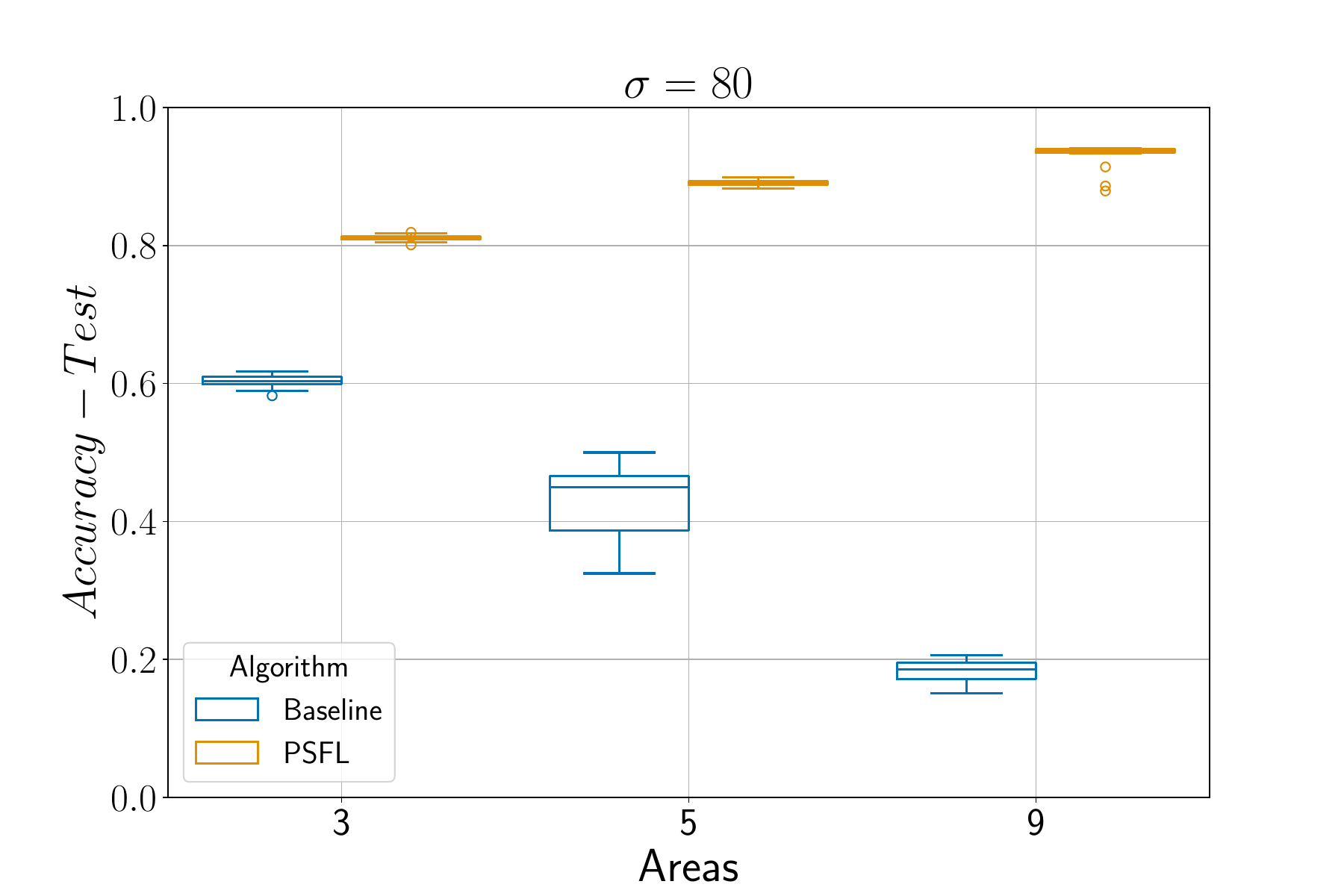}
    \end{subfigure}
    \caption{Box plots of the test accuracy for the different configurations, varying areas and the loss threshold. In \approach{}, we average the accuracy of the nodes in the same federation, whereas in the centralized FL, we average the accuracy of all nodes. We can see that
    \approach{} always outperforms the centralized FL solution. }\label{fig:test-performance-study}
\end{figure*}

The results of our study were systematically collected during two distinct phases:
\begin{enumerate}
    \item \emph{training phase:} this phase involved monitoring the variations in model performance over time and assessing the accuracy of federations (see ~\Cref{fig:training-performance-study});
    \item \emph{testing phase:} stable federations were tested by introducing new test data to evaluate the overall effectiveness (\Cref{fig:test-performance-study}).
\end{enumerate}
Analysis shows that the threshold parameter is crucial in the formation of federations. 
 A lower threshold increases the number of federations due to a more restrictive error metric, 
 particularly noticeable in scenarios with fewer areas (e.g., 3). 
 This is because the areas cover a larger portion of space and the federations depend on the error accumulated by the leader. 
 The greater the spatial coverage, the longer the path that the leader and the nodes will cover, 
 resulting in a larger \(G\).
Conversely, higher threshold values reduce the number of federations but improve their accuracy. 
 This trend is more pronounced with a larger number of areas (e.g., 9).
This is because, this time the areas are smaller and the federations, with a very large error, manage to incorporate others.

Throughout the learning process, 
 there is generally a reduction in loss and an improvement in accuracy. 
 However, under conditions of high threshold values and many areas, 
 the performance is more unstable as federations are inaccurately formed, 
 leading to ineffective learning processes and inferior models.
Despite these issues, 
 our results consistently outperform the baseline across all conditions (see \Cref{fig:simulation-snapshots} for a visual representation of the federations formed), 
 notably in configurations with more areas. 
 This is expected because, as illustrated in~\Cref{sec:background}, averaging models trained on non-IID datasets can lead to a significant reduction in 
 overall performance. 
 In this case, as the number of areas grows, the skewness of the data also increases, 
 thereby deteriorating the performance of the model 
 trained with the centralized algorithm.

\begin{figure*}
    \centering
    \begin{subfigure}[b]{0.32\textwidth}
        \includegraphics[width=\linewidth]{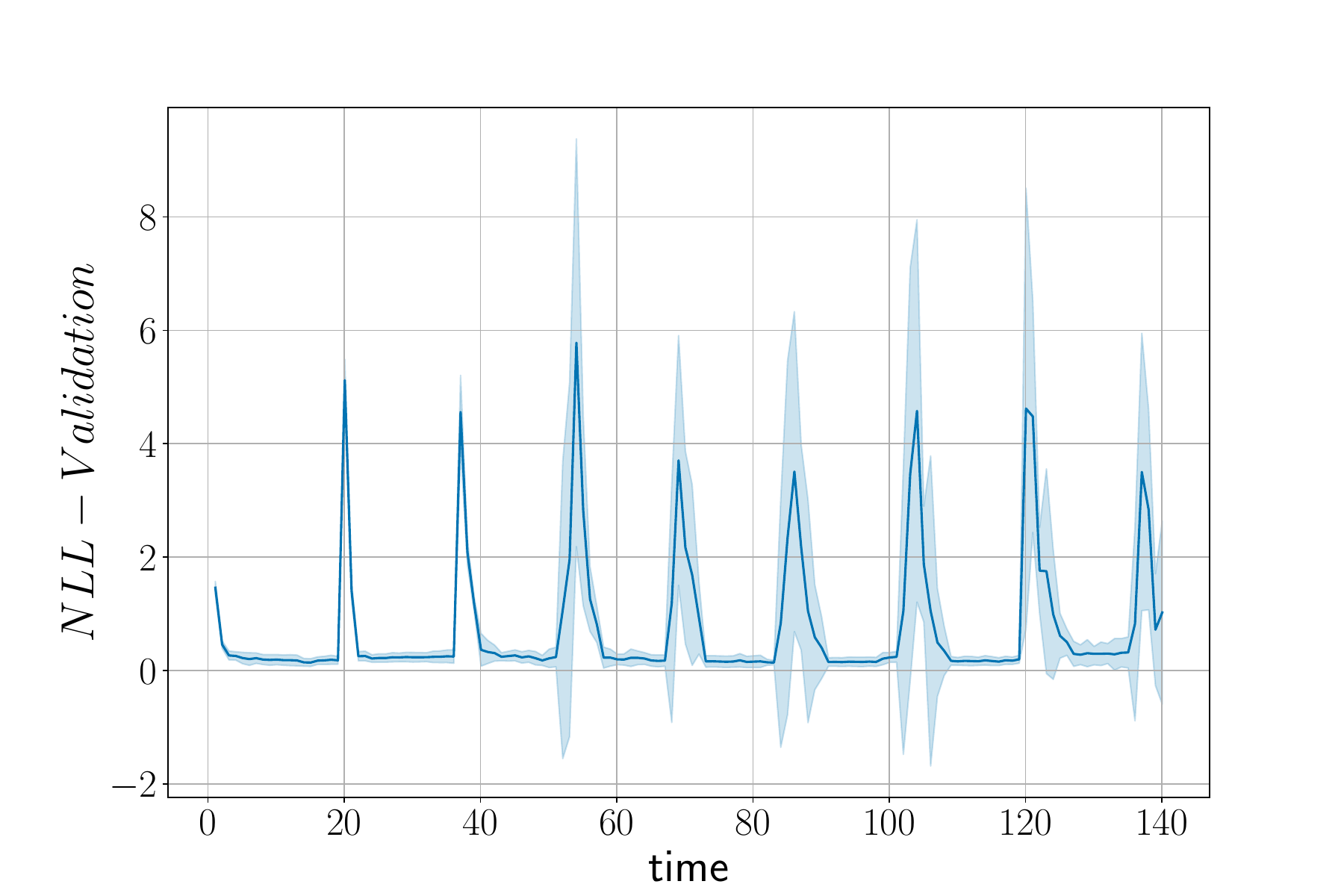}
    \end{subfigure}
    \begin{subfigure}[b]{0.32\textwidth}
        \includegraphics[width=\linewidth]{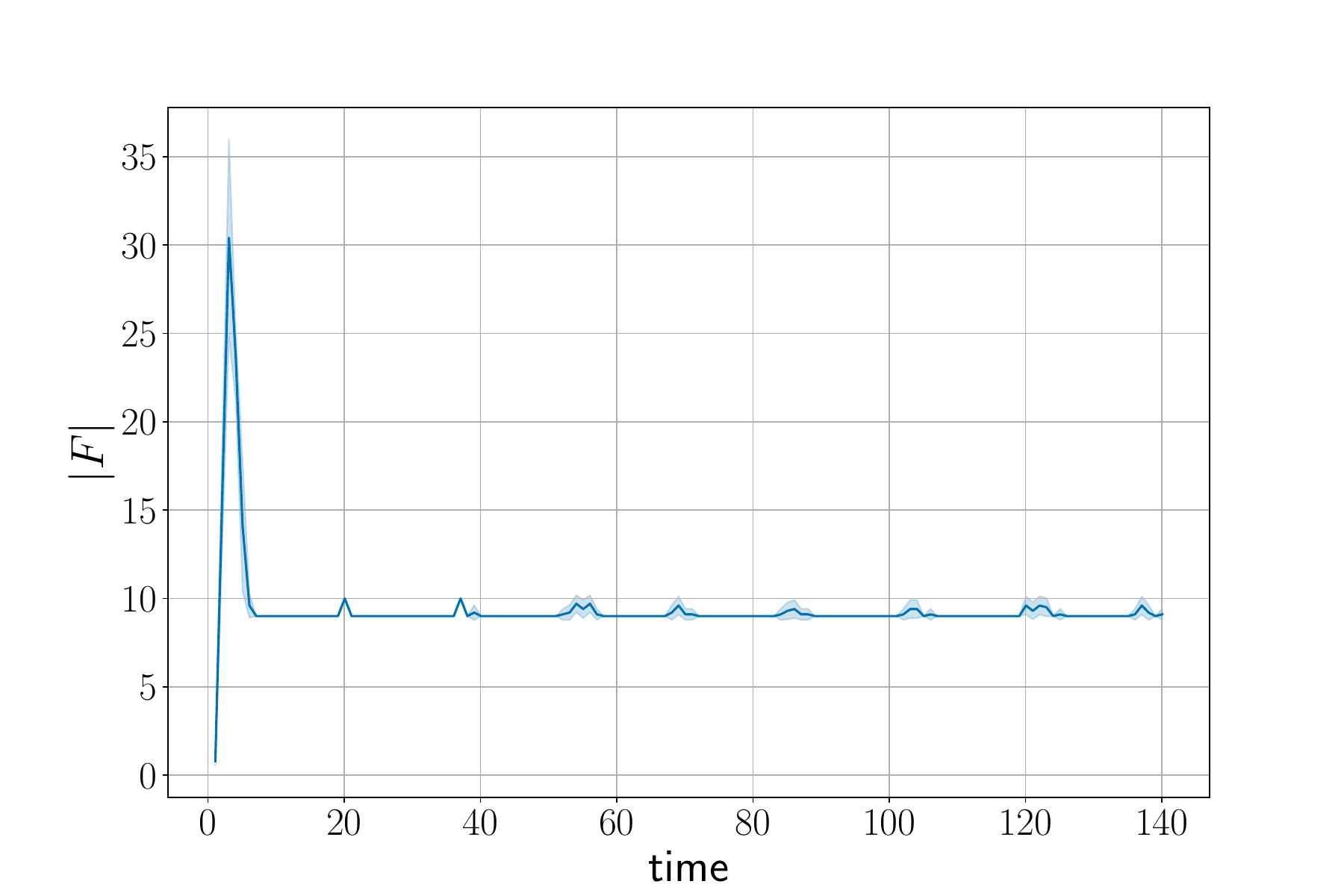}
    \end{subfigure}
    \begin{subfigure}[b]{0.32\textwidth}
        \includegraphics[width=\linewidth]{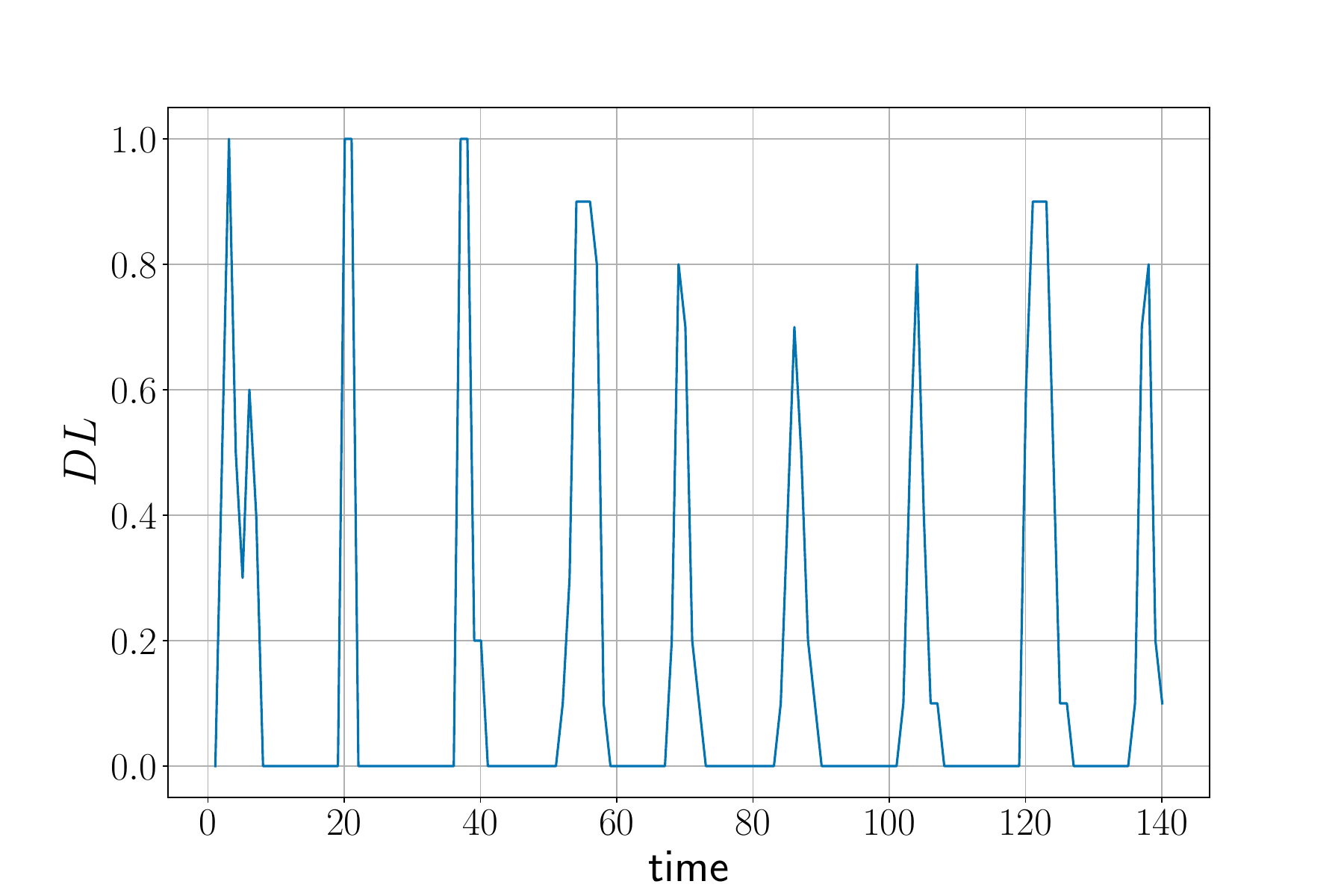}
    \end{subfigure}
    \caption{
        Plots for experiments with $\bigstar$ movement over time. 
        The results are the average of $10$ simulations with different starting seeds and $\sigma = 20.0$.
        Despite transient phases during the movement, the system reorganizes itself into a new stable configuration as the leader of the federation remains unchanged, the validation loss decreases to low values, and the number of federations remains stable.
    }\label{fig:test-movement}
\end{figure*}

Chart analysis of node movement (\Cref{fig:test-movement}) indicates that federations remain stable even as new nodes enter the system. 
 After an initial period of instability, the system reorganizes, 
 forming new federations as coherent as the previous ones.
The behaviour of the system is evident when observing two key metrics: $DL$ and $NLL - Validation$. 
Immediately following the node's movement phase, the $DL$ metric stabilizes near zero. 
Conversely, the $NLL - Validation$ initially peaks but subsequently stabilizes at a low value.
Interestingly, 
when the node transitions to a new area, it attempts to establish a new federation but is prevented by a high error metric. 
Eventually, the node integrates into an existing federation, aiding in the stabilization of the system as the number of federations, 
denoted by $|F|$, returns to the optimal count of nine.

%\meta{section length 2 pages}
\section{Conclusions and Future Work}\label{sec:conclusion}
%\meta{section length half page}
Our study introduce \approach, a novel self-organising federated learning framework
 that leverages decentralized federation of models based on geographic proximity and data similarity. 
 This approach addresses the challenges posed by data heterogeneity in traditional federated learning systems 
 by allowing for dynamic federation formation without a central coordinating node.
The results show that \approach{} %consistently 
outperforms the baseline FedAVG algorithm across various scenarios, 
particularly in configurations with a larger number of areas. 
 This highlights the potential of \approach{} to enhance federated learning processes in real-world applications, 
 where data distribution is inherently non-IID
 and when the aggregators nodes are not known a priori.

Future research may expand the evaluation of our approach to other well-known datasets, employing new partitioning methodologies to simulate 
 diverse distributions and further increasing the number of areas and nodes to encompass scenarios of varying complexities. 
Additionally, %, while we have assessed the stability of federations when a single node moves, 
 it would be interesting to evaluate the system's performance when multiple nodes move simultaneously. 
 This investigation could determine whether such movement is beneficial, 
 perhaps due to cross-fertilization among federations, 
 or detrimental due to the instability it causes in federations. 
 Building on this point, and considering the dynamic nature of the systems we are studying, 
 another area for future work could involve integrating continuous learning techniques to adapt models to data that evolves over time.

\bibliographystyle{IEEEtran}
\bibliography{IEEEexample}

\end{document}